\documentclass[lettersize,journal]{IEEEtran}
\usepackage{amsmath,amsfonts}
\usepackage{algorithmic}
\usepackage{algorithm}
\usepackage{array}
\usepackage[caption=false,font=normalsize,labelfont=sf,textfont=sf]{subfig}
\usepackage{textcomp}
\usepackage{stfloats}
\usepackage{url}
\usepackage{verbatim}
\usepackage{graphicx}
\usepackage{cite}
\usepackage{booktabs}
\usepackage{multirow}
\usepackage[table]{xcolor}

\begin{document}

\title{RoamScene3D: Immersive Text-to-3D Scene Generation via Adaptive Object-aware Roaming}

\author{Jisheng Chu, Wenrui Li,~\IEEEmembership{Member,~IEEE}, Rui Zhao, Wangmeng Zuo,~\IEEEmembership{Senior Member,~IEEE}, Shifeng Chen,\\ Xiaopeng Fan,~\IEEEmembership{Senior Member,~IEEE}
\thanks{This work was supported in part by the National Key R\&D Program of China (2023YFA1008500), the National Natural Science Foundation of China (NSFC) under grants 624B2049, 62402138, and the Fundamental Research Funds for the Central Universities under grants HIT.DZJJ.2024025. (Corresponding author: Xiaopeng Fan)}
\thanks{Jisheng Chu is with the Faculty of Computing, Harbin Institute of Technology, Harbin, China, and the Shenzhen Institutes of Advanced Technology, Chinese Academy of Sciences, Shenzhen, China. (e-mail: jschu@stu.hit.edu.cn)}
\thanks{Wenrui Li, and Wangmeng Zuo are with the Faculty of Computing, Harbin Institute of Technology, Harbin, China. (e-mail: liwr@hit.edu.cn; wmzuo@hit.edu.cn).}
\thanks{Rui Zhao is with the Nanyang Technological University, Singapore. (e-mail: ruizhao26@gmail.com)}
\thanks{Shifeng Chen is with the Shenzhen Institutes of Advanced Technology, Chinese Academy of Sciences, Shenzhen 518172, China (e-mail: shifeng.chen@siat.ac.cn)}
\thanks{Xiaopeng Fan is with the Faculty of Computing, Harbin Institute of Technology, Harbin, China, the Peng Cheng Laboratory, Shenzhen, China, and also with Harbin Institute of Technology Suzhou Research Institute, Suzhou, China. (e-mail: fxp@hit.edu.cn)}
}



\maketitle

\begin{abstract}
% Generating immersive 3D scenes from texts is a core task in computer vision, crucial for applications in virtual reality and game development. Existing methods typically follow predefined or random trajectories, iteratively inferring occluded content from novel viewpoints. However, they face challenges to fully exploit inner relationships among salient objects, thus roaming cameras are unable to explore the scene adaptively. Moreover, current inpainting models struggle to fill holes due to complex occlusions under novel views plausibly. To address these challenges, we propose a novel text-to-3D generation framework, RoamScene3D, that produces consistent and photorealistic scenes. It autonomously explores scenes by reasoning about spatial relations among objects. Specifically, our method generates an RGBD panorama from a text prompt to initialize a 3D scene. A vision-language model (VLM) is then employed to infer a scene graph that encodes object relations. By perceiving the boundaries of the salient objects in the graph, our approach constructs a closed roaming trajectory. At each viewpoint, the camera applies adaptive positional offsets aware of the nearest object. We iteratively generate novel panoramic views and create the scene with 3D Gaussian Splatting. To make the RGBD inpainting model adaptive to camera motion, we curated a synthetic panoramic dataset integrating authentic camera trajectories to fine-tune the inpainting model. Extensive experiments clearly demonstrate the superiority of our proposed method over state-of-the-art approaches.

Generating immersive 3D scenes from texts is a core task in computer vision, crucial for applications in virtual reality and game development. Despite the promise of leveraging 2D diffusion priors, existing methods suffer from spatial blindness and rely on predefined trajectories that fail to exploit the inner relationships among salient objects. Consequently, these approaches are unable to comprehend the semantic layout, preventing them from exploring the scene adaptively to infer occluded content. Moreover, current inpainting models operate in 2D image space, struggling to plausibly fill holes caused by camera motion. To address these limitations, we propose RoamScene3D, a novel framework that bridges the gap between semantic guidance and spatial generation. Our method reasons about the semantic relations among objects and produces consistent and photorealistic scenes. Specifically, we employ a vision-language model (VLM) to construct a scene graph that encodes object relations, guiding the camera to perceive salient object boundaries and plan an adaptive roaming trajectory. Furthermore, to mitigate the limitations of static 2D priors, we introduce a Motion-Injected Inpainting model that is fine-tuned on a synthetic panoramic dataset integrating authentic camera trajectories, making it adaptive to camera motion. Extensive experiments demonstrate that with semantic reasoning and geometric constraints, our method significantly outperforms state-of-the-art approaches in producing consistent and photorealistic scenes. Our code is available at https://github.com/JS-CHU/RoamScene3D.

\end{abstract}


\begin{figure}[t]
\centering
\includegraphics[width=0.999\columnwidth]{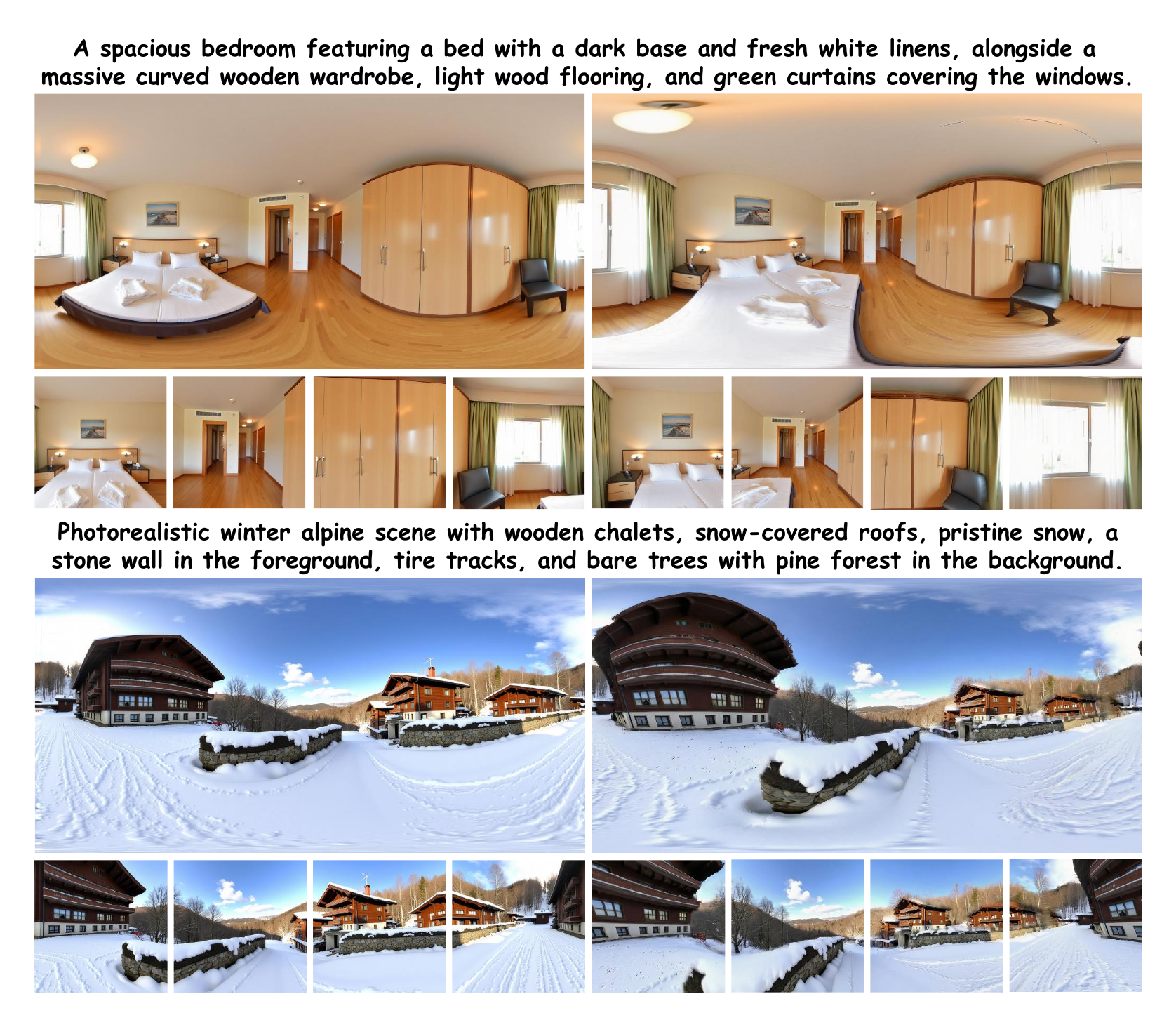} 
\caption{The visual results of the proposed method. Given a text description, our method is capable of generating arbitrary indoor and outdoor scenes. The synthesized environments are photorealistic and immersive, enabling navigable walkthroughs. The figure illustrates a trajectory transitioning from one viewpoint to another within the scene.}
\label{figure:preview}
\end{figure}

\section{Introduction}
% \IEEEPARstart{T}{he} objective of text-to-3D scene generation is to produce immersive 3D environments conditioned on natural language input, with a strong emphasis on maintaining spatial coherence and geometric fidelity. Driven by recent advances in both 2D image generation techniques \cite{nichol2021improved,rombach2022high} and 3D representation learning \cite{mildenhall2021nerf,3dgs}, this task has attracted increasing attention from the research community. Text-to-3D scene generation leverages powerful pretrained image generation models to derive 2D priors from text descriptions, which are subsequently integrated with textual information to generate 3D scene representations. It holds significant promise for a wide range of downstream applications, including virtual reality and game development.

\IEEEPARstart{A}{UTOMATED} 3D content creation is a pivotal research area at the intersection of computer vision and computer graphics \cite{mildenhall2021nerf,3dgs,bautista2022gaudi,chen2023scenedreamer}. 
Within this domain, while recent works excel in specific assets \cite{yu2025food3d,zhu2025tg}, text-to-3D scene generation aims to synthesize immersive, spatially coherent 3D environments from natural language descriptions. 
This task has garnered varying degrees of attention due to the exponential growth of the digital ecosystem and the increasing complexity of virtual applications \cite{li2026hyperbolic,hollein2023text2room,fridman2023scenescape,li2025riemann,chu2025digging,li2025hyperbolic}.
The rapid evolution of the Metaverse, Virtual Reality (VR), and Augmented Reality (AR) has precipitated an unprecedented demand for high-fidelity 3D assets.
These staggering requirements underscore the limitations of traditional manual modeling pipelines, highlighting the urgent necessity for efficient, intelligent generation techniques to streamline the production of massive 3D environments \cite{hu2024scenecraft,yu2025wonderworld}.
A dominant paradigm in recent research is to use the rich visual priors encapsulated in powerful pretrained 2D text-to-image diffusion models \cite{nichol2021improved,rombach2022high}. Leveraging these 2D generative priors, current methods enable the synthesis of complex textures and structures.

Existing methods typically adopt a strategy that iteratively completes a scene, relying on a recurrent cycle involving unprojection, rendering, and inpainting.  Pioneering approaches \cite{hollein2023text2room,lucid,fridman2023scenescape} have demonstrated the feasibility of generating semantically grounded 3D scenes by leveraging 2D diffusion models. They initialize a scene through a combination of image generation and depth estimation, and subsequently place a series of novel cameras in 3D space for iterative scene construction. However, they often operate under predefined camera trajectories and perspective views, which constrain the flexibility of the generated scenes and prevent comprehensive spatial coverage. Several subsequent efforts \cite{li2024scenedreamer360,zhou2024dreamscene360,pu2024pano2room} have turned to panoramic generation paradigms. They employ panoramic views for this process, enabling the synthesis of 360\textdegree\ views. Although these methods improve the spatial coverage of the scene, their camera trajectories still remain either manually predefined or randomly generated, constrained to restricted spatial regions without the ability for adaptive movement. For example, DreamScene360 \cite{zhou2024dreamscene360} applies random camera perturbations in three distinct stages near the coordinate origin, while Pano2Room \cite{pu2024pano2room} follows a predefined, fixed camera motion trajectory. Furthermore, the generative process lacks awareness of object spatial relationships and the semantic structure of the scene. Inspired by scene composition techniques, recent approaches have begun to incorporate scene structure into the generation process. SceneCraft \cite{hu2024scenecraft} utilizes an explicit layout to control scene arrangement, but its reliance on manually designed inputs limits flexibility. ScenePainter \cite{xia2025scenepainter} extracts scene graphs from text and images to model object relationships, though it does not leverage them for camera trajectory planning. LayerPano3D \cite{yang2025layerpano3d} mitigates spatial occlusion via depth based layering, yet lacks understanding of semantic object relationships. Beyond the issues discussed above, existing image inpainting models exhibit limitations in realistically completing missing regions caused by intricate occlusions from novel viewpoints.

In summary, current text-to-3D scene generation approaches face the following challenges:
\textbf{1) Trajectory Adaptivity:} Manually defined or randomly perturbed camera trajectories lead to suboptimal viewpoint coverage and insufficient focus on salient objects.
\textbf{2) Objects Relation Reasoning:} Object relation reasoning remains underexplored in text-to-3D scene generation, limiting the model’s ability to capture spatial and semantic dependencies among objects.
\textbf{3) Motion-injected Inpainting:} Maintaining geometric consistency across rendered novel views remains a key challenge, particularly under large parallax and complex occlusions. Existing RGBD inpainting models are not inherently adaptive to camera motion, leading to spatial distortions and appearance inconsistencies.

To address these challenges, we propose RoamScene3D, a novel text-to-3D generation framework that unifies semantic scene understanding, adaptive trajectory planning, and motion-injected inpainting to construct spatially coherent, photorealistic 3D environments. Our approach departs from conventional static or random camera path paradigms by dynamically guiding the scene exploration process using semantic object reasoning. Specifically, our pipeline begins by generating an initial RGBD panorama conditioned on a text prompt using a pretrained diffusion model paired with a panorama depth estimation module. This serves as the foundational view of the 3D scene. We then employ a VLM to extract a high level scene graph, which encodes semantic relationships. This graph allows us to identify clusters of salient objects, which serve as anchor points for the virtual camera's navigation. To determine a meaningful and spatially comprehensive trajectory, we introduce a closed roaming strategy. By disentangling the most salient objects and estimating their boundaries in 3D space, we sample a series of viewpoints that maintain proximity to object surfaces while adapting to local scene geometry. Each camera pose incorporates adaptive positional offsets with respect to nearby object boundaries, promoting a more immersive and detailed traversal of the scene. At each novel viewpoint, we render the expected panoramic view and invoke a motion-injected RGBD inpainting model, fine-tuned on a scene dataset collected from the Matterport3D environment. This dataset comprises pairs of panoramas with annotated camera transformations, enabling the model to learn motion conditioned inpainting priors that enhance consistency under large viewpoint shifts and complex occlusions. The sequence of rendered panoramas is then integrated into a unified representation using 3D Gaussian Splatting, allowing for efficient and photorealistic scene reconstruction. Through this integrated framework, RoamScene3D not only ensures adaptive spatial coverage of the scene but also maintains semantic awareness and geometric consistency across views. The contributions can be summarized as follows:
\begin{itemize}
    \item We propose RoamScene3D, a novel framework for immersive 3D scene generation that integrates semantic understanding, adaptive exploration, and panoramic scene construction. Our method produces photorealistic and spatially consistent scenes from text prompts.
    \item We introduce an adaptive roaming trajectory planning module that leverages object relation reasoning to guide viewpoint selection. By constructing a scene graph, our method identifies semantic relationships among objects and adaptively plans a closed trajectory that ensures both comprehensive spatial coverage and semantic focus.
    \item We propose a motion-injected panorama inpainting model, fine-tuned on a curated panorama dataset with annotated camera motions. The fine-tuned model effectively adapts to occlusions induced by viewpoint transitions, enabling spatially coherent and visually plausible completions under large parallax and complex scene geometry.
\end{itemize}

Our method demonstrates clear performance gains in terms of visual quality and object coherence compared to state-of-the-art baselines. The rest of the paper is organized as followed: Section II provides the research background on 3D scene generation. Section III offers an detailed illustration of the proposed RoamScene3D. Section IV provides evidence of the model’s effectiveness through a series of experiments and visual comparisons. In conclusion, Section V highlights the main findings and contributions of this work.

\section{Related Work}

\subsection{3D Scene Generation Methods}
The focus of 3D generation research has evolved from object generation to holistic scene generation, which takes textual or visual prompts as input and generates an immersive 3D scene. These approaches can be broadly categorized into feedforward generation \cite{bautista2022gaudi,chen2023scenedreamer,Kim_2023_CVPR,liu2023pyramid,wu2024blockfusion,meng2025lt3sd,Ren_2022_CVPR,Li_2023_CVPR} and optimization-based generation~\cite{hollein2023text2room,lucid,zhang2024text2nerf,Yu_2024_CVPR,li2024scenedreamer360,zhou2024dreamscene360,pu2024pano2room,Wang_2024_perf,fridman2023scenescape,yang2025layerpano3d,xia2025scenepainter,RGBD2}. Feedforward methods aim to directly produce 3D scene representations by training on 3D datasets. Early works, such as GAUDI~\cite{bautista2022gaudi} and SceneDreamer~\cite{chen2023scenedreamer}, utilized generative models to capture scene distributions. More recently, recurrent diffusion frameworks have been proposed to infer 3D point cloud structures directly from single images, significantly enhancing geometric details \cite{zhou2025recurrent}. To better handle complex viewpoints, \cite{Ren_2022_CVPR} incorporates a camera bias into an autoregressive model to generate consistent scene images along camera trajectories. More recent approaches have explored efficient representations, such as triplane extrapolations~\cite{wu2024blockfusion,xie2025tri}, hierarchical latent trees~\cite{meng2025lt3sd}, and multiscale generative frameworks using radiance fields based on grids~\cite{Li_2023_CVPR,zuo2025learning,li2025ustc,chen2025MuTri}, to improve generation quality and diversity. However, these methods are constrained by the scarcity of 3D scene datasets on a large scale, which often results in generated scenes lacking realistic texture details.

To overcome data limitations, optimization-based methods leverage powerful 2D priors from image generation models to guide 3D optimization. This paradigm typically follows an iterative process of rendering and refining. Methods such as Text2Room~\cite{hollein2023text2room}, SceneScape~\cite{fridman2023scenescape}, and Text2NeRF~\cite{zhang2024text2nerf} generate initial views from text and progressively inpaint missing regions to construct consistent 3D meshes or NeRF representations. Similarly, Cai et al.~\cite{Cai_2023_ICCV} simulate flythrough trajectories using a conditional diffusion model for concurrent inpainting and refinement, while RoomDreamer~\cite{RoomDreamer} utilizes a cubemap representation to fill uncovered regions via outpainting for mesh optimization. LucidDreamer~\cite{lucid} further improves this by unprojecting multiview images generated by Stable Diffusion onto a point cloud for optimization. Building on approaches based on points, Yu et al.~\cite{Yu_2024_CVPR} combine descriptions guided by LLMs with refined depth for point cloud fusion, later introducing Fast Layered Gaussian Surfels (FLAGS) for iterative control~\cite{yu2025wonderworld}. Zhang et al.~\cite{zhangmonst3r} extend optimization to video processing by using point maps to refine global point clouds. To ensure deeper consistency, Zhang et al.~\cite{zhang20243d} and others \cite{chen2025learning,ju2025revisiting,xiao2025enhanced} lift 2D feature maps into 3D space to enforce consistency between views inpainted by DIBR and volumetrically rendered views, while Li et al.~\cite{li2024art3d} tackle semantic misalignment in stylized generation by aligning internal diffusion features. To enhance spatial coverage, recent works have extended this pipeline to panoramic generation. DreamScene360~\cite{zhou2024dreamscene360} and SceneDreamer360~\cite{li2024scenedreamer360} generate panoramic images to initialize 3D Gaussian Splatting, while Pano2Room~\cite{pu2024pano2room} employs panoramic depth estimation and RGBD inpainting.

Despite their ability to generate complex and visually realistic environments, these overall generation methods generally do not explicitly model geometric relationships between objects. The quality of individual objects is often compromised, and the lack of semantic awareness prevents scenes with consistent object placements. Furthermore, ensuring multiview consistency across large viewpoints remains challenging. We note that similar challenges in 3D semantic scene completion have been addressed by leveraging sparse guidance \cite{mei2024camera} and enhancing geometry-semantic features \cite{xiao2025enhanced}, which underscores the importance of geometric constraints in 3D tasks.

\subsection{Semantic Structure Guidance}
A straightforward approach to incorporating semantics is Scene Composition \cite{wang2019planit,paschalidou2021atiss,gao2024graphdreamer,zhai2023commonscenes,tang2024diffuscene,yang2024physcene,fang2023ctrl,xu2023discoscene,zhou2024gala3d}, which constructs scenes by assembling objects based on layout rules. Following early probabilistic and grammar models \cite{merrell2011interactive,fisher2012example,qi2018human}, deep autoregressive approaches \cite{wang2019planit,paschalidou2021atiss} have evolved to handle scenarios ranging from hybrid interior synthesis \cite{zhao2024roomdesigner} to unbounded city scales \cite{Xie_2024_CVPR}. To capture complex interactions, recent works leverage semantic scene graphs and graph representation learning \cite{gao2024graphdreamer,zhai2023commonscenes,li2025language,InstructScene,li2025spiking}, while diffusion-based methods \cite{tang2024diffuscene,yang2024physcene,fang2023ctrl,wei2023lego} introduce physical constraints and rearrangement capabilities for comprehensive layout generation. Beyond explicit layouts, optimization-based strategies integrate guidance into implicit representations (e.g., NeRF or 3DGS) to enhance manageability. These approaches have been extended to enable object-centric control via spatial disentanglement \cite{zhang2024towards,epstein2024disentangled}, ensure volumetric consistency through feature grids \cite{Wu_2023_ICCV,Bahmani_2023_ICCV}, and refine texture quality using panoramic or proxy-based constraints \cite{bai2024360,Schult_2024_CVPR,jiao2025clip}. Despite these advances, assembling independent assets often yields ``toylike'' results lacking global illumination and realistic consistency.

To combine the structural control of composition with the visual fidelity of overall generation, recent research has sought to integrate semantic descriptors (layouts or scene graphs) directly into the generative pipeline. This strategy borrows the idea of decoupling structure from appearance to guide the synthesis process. For instance, SceneCraft~\cite{hu2024scenecraft} leverages Large Language Models (LLMs) to generate explicit layout code. However, its reliance on manually designed inputs or rigid code structures limits the flexibility and diversity of the generated scenes. ScenePainter~\cite{xia2025scenepainter} advances this by extracting scene graphs from text and images, ensuring better semantic alignment. Nevertheless, it focuses primarily on static consistency and does not explore the scene adaptively. Similarly, LayerPano3D~\cite{yang2025layerpano3d} decomposes the scene into layers based on depth. However, it lacks a deep semantic understanding of the logical relationships between objects.

In summary, current scene generation methods produce photorealistic but semantically unaware environments, resulting in geometric inconsistencies for complex objects. Conversely, scene composition methods offer strong semantic control via layouts and scene graphs but are typically restricted to simple, assembled scenes. Although recent hybrid approaches improve coherence via semantic guidance, they remain confined to static composition and fail to leverage semantics for adaptive scene exploration. Their reliance on passive trajectories consequently restricts the adaptive observation of salient objects. Unlike prior works, our proposed \textbf{RoamScene3D} bridges this gap by integrating semantic reasoning into the generation pipeline. Our approach facilitates adaptive exploration of salient objects, ensuring comprehensive spatial coverage. This allows for high-fidelity generation that remains geometrically consistent even when facing the significant parallax and complex occlusions inherent in adaptive roaming.

\section{Methodology}
\begin{figure*}[t]
\centering
\includegraphics[width=0.99\textwidth]{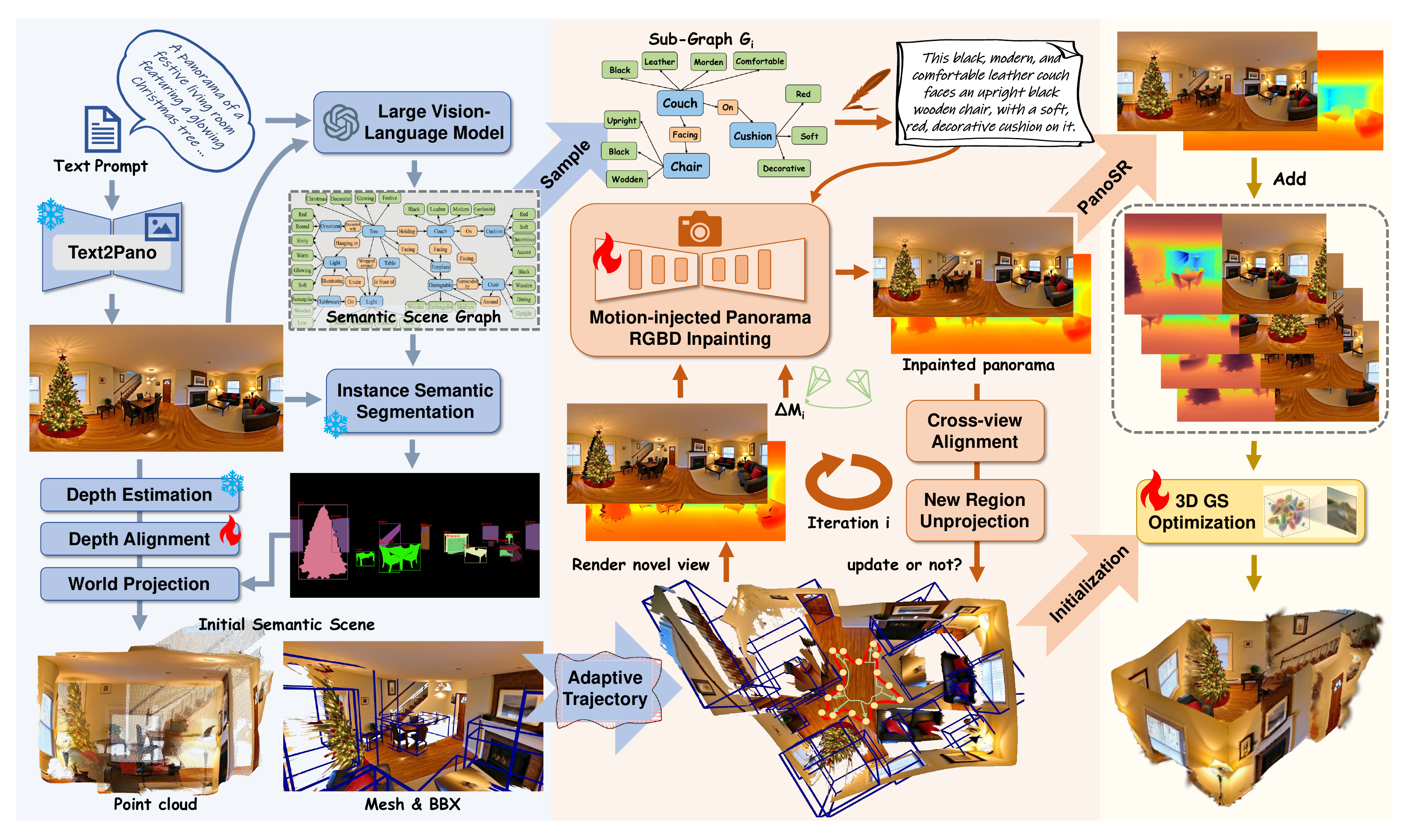}
\caption{Overview of the RoamScene3D framework. Given a text prompt, our method first initializes the scene by generating an RGBD panorama and unprojecting it into a coarse 3D representation. To ensure semantic fidelity, we utilize a VLM to construct a Semantic Scene Graph, which encodes object relationships and guides to adaptively generate a closed, object-aware camera trajectory. Subsequently, we employ a Motion-injected Panorama RGBD Inpainting model to synthesize consistent novel views conditioned on camera movement, effectively handling disocclusions. Finally, the sequence of rendered panoramas is integrated via 3D Gaussian Splatting optimization to produce a photorealistic and spatially coherent immersive 3D scene.}
\label{figure:structure}
\end{figure*}
% The architecture of RoamScene3D is illustrated in Fig.\ref{figure:structure}, comprising four key modules: Scene Initialization, Roaming Trajectory Planning, Motion-injected RGBD Panorama Inpainting, and Scene Optimization. 
The architecture of RoamScene3D is illustrated in Fig.\ref{figure:structure}, taking a text prompt as input and generating a 360\textdegree\ immersive scene represented by 3D Gaussian Splatting. To ensure adaptive and object-aware exploration, we introduce a Roaming Trajectory Planning module that constructs a closed trajectory guided by semantic object relationships extracted from a VLM. At each sampled viewpoint, our Motion-injected Inpainting module fills missing content conditioned on camera motion, enhancing geometric consistency under occlusion and parallax. The resulting panoramas are fused using 3D Gaussian Splatting in the Scene Optimization stage to produce a coherent and photorealistic 3D environment.
% RoamScene3D (Fig.\ref{figure:structure}) generates immersive 360\textdegree\ 3D Gaussian Splatting scenes from text prompts through four modules. Roaming Trajectory Planning utilizes VLM-derived semantic relationships for adaptive exploration, while Motion-injected Inpainting ensures geometric consistency by conditioning content generation on camera motion. Finally, Scene Optimization fuses the resulting panoramas into a coherent photorealistic environment.
\subsection{Scene Initialization}
\label{sec:initialization}
% Due to the scarcity of high-quality 3D scene datasets and the inherent complexity of 3D environments, directly generating photorealistic 3D scenes from text remains highly challenging. Consequently, current approaches often leverage powerful text-to-image models to initialize 3D scenes using 2D priors.
From text descriptions, we utilize a pretrained panorama generation model following \cite{yang2025layerpano3d} to generate the initial 360\textdegree\ panorama $\mathbf{I}_0$ of resolution $H\times W$. According to the initial panorama, we unproject the image into 3D space. Inspired by previous methods \cite{hollein2023text2room,pu2024pano2room}, we get a colored point cloud and then connect these points to construct the scene mesh.

The generated panorama is projected onto a regular icosahedron, resulting in 20 perspective tangent images with overlaps. We employ pretrained depth and normal estimators to predict the geometry of the perspective images and subsequently align them to get the panoramic distance map. To achieve global geometric coherence, we fit a differentiable sphere distance field $f_\varphi:\mathbb{S}^2\!\rightarrow\mathbb{R}$ that maps each viewing direction $\mathbf{d}$ to a distance $r=f_\varphi(\mathbf{d})$.
We optimize the field by jointly optimizing global scales $\{\alpha_k\}$ for all views and spatially varying local biases $b_k^d(u,v)$ and $\mathbf{b}_k^n(u,v)$. By constructing an orthonormal tangent frame $(\mathbf{o}_a, \mathbf{o}_b)$ such that $(\mathbf{o}_a, \mathbf{o}_b) \perp \mathbf{d}$, we derive two tangent vectors $\mathbf{v}_a$ and $\mathbf{v}_b$ orthogonal to the surface normal to enforce joint normal constraints:
\begin{equation}
\mathbf{v}_a=\frac{(\nabla f_\varphi\!\cdot\!\mathbf{o}_a)\,\mathbf{d}+\mathbf{o}_a}{\|(\nabla f_\varphi\!\cdot\!\mathbf{o}_a)\,\mathbf{d}+\mathbf{o}_a\|},\quad
\mathbf{v}_b=\frac{(\nabla f_\varphi\!\cdot\!\mathbf{o}_b)\,\mathbf{d}+\mathbf{o}_b}{\|(\nabla f_\varphi\!\cdot\!\mathbf{o}_b)\,\mathbf{d}+\mathbf{o}_b\|}.
\end{equation}
% \begin{equation}
% \mathbf{n}=\frac{\mathbf{v}_a\times\mathbf{v}_b}{\|\mathbf{v}_a\times\mathbf{v}_b\|}.
% \end{equation} 

The objective combines distance $\mathcal{L}_d$, normal alignment $\mathcal{L}_n$, scale regularization $\mathcal{L}_{reg}$, and total variation $\mathcal{L}_{tv}$ terms:
\begin{equation}
\begin{aligned}
\min_{\substack{\varphi,\{\alpha_k\}, \{b_k^d\},\{\mathbf{b}_k^n\}}}\;
\sum_k \Big( \mathcal{L}_d + \lambda_n\mathcal{L}_n + \lambda_{reg}\mathcal{L}_{reg} + \lambda_{tv}\mathcal{L}_{tv} \Big),
\end{aligned}
\end{equation}
\begin{equation}
\mathcal{L}_d
= \operatorname{SmoothL1}\!\Big(
    \alpha_k\,\tilde{r}_k(u,v)
    + b_k^d(u,v),
    \ f_\varphi(\mathbf{d}_k(u,v))
\Big),
\label{eq:ld}
\end{equation}
\begin{equation}
\mathcal{L}_n
= \operatorname{SmoothL1}\!\left(
    \left\{
    \begin{aligned}
        &\langle \mathbf{v}_a,\ \mathbf{n}_k(u,v) + \mathbf{b}_k^n(u,v) \rangle, \\
        &\langle \mathbf{v}_b,\ \mathbf{n}_k(u,v) + \mathbf{b}_k^n(u,v) \rangle
    \end{aligned}
    \right\},
    0
\right),
\label{eq:ln}
\end{equation}
\begin{equation}
\mathcal{L}_{reg} = \|\mathrm{softplus}(\alpha_k)-1\|_2^2,
\mathcal{L}_{tv} = f_{TV}(b_k^d)+f_{TV}(\mathbf{b}_k^n),
\end{equation}
where $\tilde{r}_k(u,v)$ and $\mathbf{n}_k(u,v)$ denote the distance map and normal vector predicted by pretrained estimators respectively. $f_{TV}(\cdot)$ smoothes the local bias map by penalizing differences between adjacent elements, thereby improving geometric consistency and optimization stability.

With the sphere distance field as the geometric foundation, we unproject the panorama into a colored point cloud to serve as the vertex set ${V}$ and the colors set ${C}$ for the scene mesh ${M} = \{{V}, {C}, {F}\}$. The position of each vertex is determined by mapping the pixel coordinates $(u, v)$ to polar angles $\phi = \pi \cdot v / H$ and $\theta = 2\pi \cdot u / W - \pi/2$ via spherical projection.
% \begin{equation}
% \begin{pmatrix} x \\ y \\ z \end{pmatrix} = r(\phi, \theta) \cdot \begin{pmatrix} \sin\phi \cos\theta \\ \cos\phi \\ \sin\phi \sin\theta \end{pmatrix}
% \end{equation}
The positive directions of the $x$, $y$, and $z$ axes of the world coordinate system are defined to point forward, upward, and rightward, relative to the panorama. The face set ${F}$ is constructed by inheriting the intrinsic grid connectivity of the original pixels. Adjacent points, which correspond to neighboring pixels in the panorama, are triangulated to form a continuous surface. To preclude ``stretched" polygons, the mesh topology is refined using a depth gradient criterion. Connectivity is selectively pruned where sharp transitions occur in the predicted distance map $r$, effectively interpreting these high gradient zones as structural boundaries.

\subsection{Adaptive Roaming Trajectory Planning}
To enable the roaming camera to explore the scene in a semantically meaningful and geometrically comprehensive manner, we propose an Adaptive Roaming Trajectory Planning module. Drawing inspiration from virtual view selection strategies \cite{mu2024learning}, the key idea is to dynamically generate a closed trajectory guided by disentangling salient objects from the scene and identifying their spatial relationships. We derive the semantic scene graph from a VLM and then analyze the positions and relationships of the most salient objects to obtain their bounding boxes. By understanding boundaries of these objects, this planning mechanism ensures that camera movements prioritize visually and semantically important regions while maintaining scene coverage and structural coherence.

\subsubsection{Scene Graph Construction and Salient Objects Disentanglement}
Given the text prompt and the generated panorama, we leverage a pretrained VLM to represent the scene as a graph ${G} = f_{VLM}(\mathbf{I}_{0}, T_{prompt})$, encoding both the constituent objects and their mutual spatial associations. The directed graph is formalized as ${G}=({O},{R},{A})$. Objects $O$ is a set of entities identified in the text and panorama, ${O}=\{o_1, o_2, \dots, o_n\}$, where each $o_i$ represents a distinct semantic instance such as \textit{sofa} or \textit{coffeetable}. For each object $o_i \in {O}$, we associate a set of visual descriptors ${A}(o_i)=\{a_{i,1}, a_{i,2}, \dots, a_{i,m}\}$. The spatial associations between objects are defined as triples of the form ${R}(o_i) \subseteq \{(o_i, r_{ij}, o_j)\}$, where $r_{ij}$ denotes a directed relationship from $o_i$ to $o_j$. For example, $(\textit{coffeetable}, \text{on}, \textit{rug})$. This structured representation maintains geometric consistency throughout the entire scene generation pipeline. However, accounting for the intricate details of every object incurs prohibitive computational overhead. To maintain an optimal trade-off between processing speed and visual fidelity, we use the VLM to select $k$ most salient objects $O_s=\{o_{s1},o_{s2},\dots,o_{sk}\}$. This allows the algorithm to prioritize the most significant scene elements.

Leveraging the scene graph and identified salient objects, our method performs instance segmentation on the panorama $\mathbf{I}_0$ to localize their corresponding pixels. A pretrained panoptic segmentation model is employed to partition all the instances. We leverage the CLIP image and text encoders to extract features for the segmented regions and the semantic categories suggested by the VLM. We then compute the cosine similarity between these features to assign category names to each segmented region, effectively annotating the pixels of each object with their corresponding semantic labels. Furthermore, these semantic labels are mapped onto the 3D point cloud via pixel-wise correspondence with the panoramic images. By computing the spatial extent of the point cloud for each identified object, 3D bounding boxes are subsequently generated. This approach facilitates the effective disentanglement of objects within the scene, enabling the model to achieve an explicit understanding of object boundaries.

\subsubsection{Adaptive Trajectory Generation}
\begin{figure}[t]
\centering
\includegraphics[width=0.999\columnwidth]{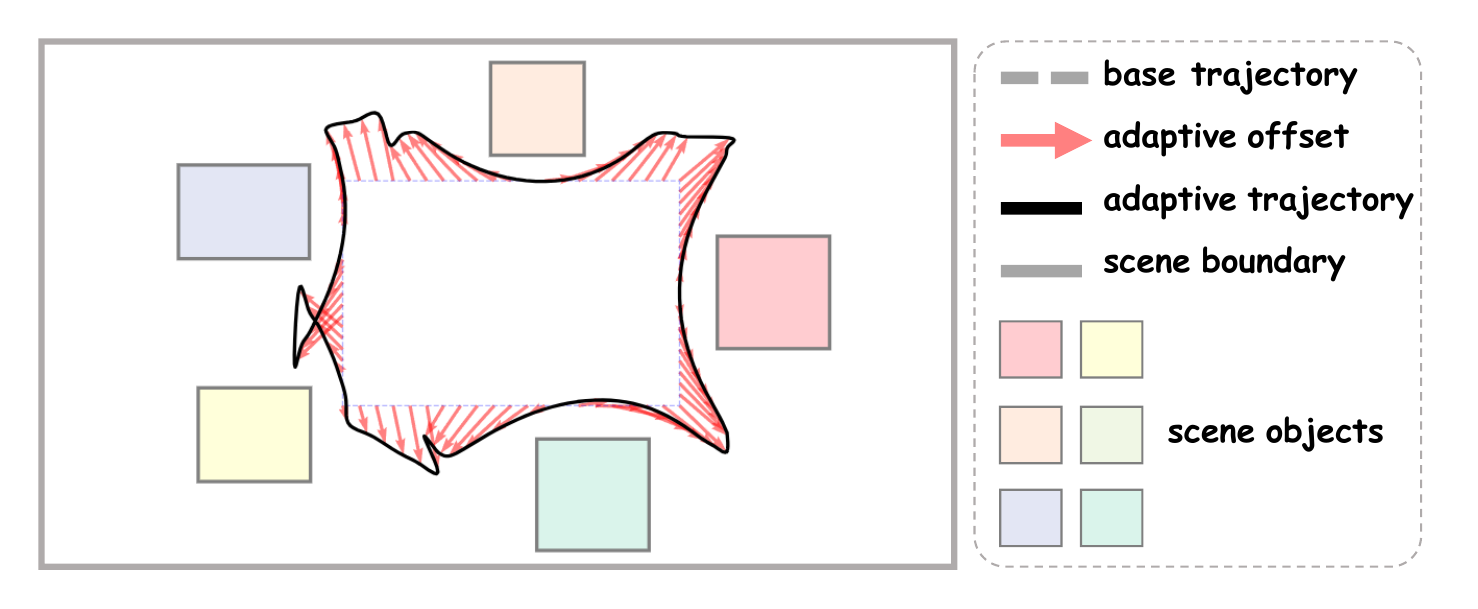} 
\caption{The visualization of the generated adaptive trajectory. It is plotted in a 2D plane for simplification. The outermost gray rectangle denotes the scene boundary, while the colored rectangles represent objects within the scene. The figure illustrates the evolution from the initial trajectory (gray dashed line) to the final trajectory (solid black line) following the applied perturbation (pink).}
\label{figure:trajectory}
\end{figure}
Based on the object bounding boxes decoupled from the scene, we construct a closed trajectory ${Tr(\phi)}$. We prioritize a continuous trajectory over random camera placement with large baselines, as this continuity is essential for maintaining consistency in novel views. Instead of using predefined or random trajectories, our method adaptively generates distinct trajectories corresponding to varying $T_{prompt}$ and $\mathbf{I}_{0}$. We first establish a collision-free base trajectory utilizing the initial panoramic distance map $D(\phi, \theta)$, where $\phi$ and $\theta$ define the azimuth angle of each pixel. We extract the distance values along the equatorial line of the panorama and scale them by a factor $\gamma \in (0,1)$ to define the initial trajectory as ${Tr}^0=\{\mathbf{p}_l\}_{l=1}^L$. ${Tr}^0$ represents a horizontal irregular closed curve centered at the scene origin, discretized into $L$ viewpoints. Each $\mathbf{p}_l$ is calcuted as:
\begin{equation}
\mathbf{p}_l = \gamma \cdot D(\phi_l,0) \cdot (\cos \phi_l, 0, \sin \phi_l), \quad \phi_l \in [0, 2\pi).
\end{equation}

% To ensure the visual integrity of salient objects, the camera should deviate from the base trajectory to observe objects from informative viewpoints. For each salient object $o_i \in O_s$, represented by its bounding box $\mathbf{B}_i$, we aim to observe it from angles that reveal its side structures. Simply pulling the camera towards the object center often leads to flat views of object faces. To maximize information gain and reveal occluded parts (e.g., sides and back), we introduce a corner-biased mechanism. The optimal observation directions for a box-shaped object are along its diagonal rays, which allow the camera to capture three adjacent faces simultaneously. Intuitively, the optimal viewpoints to "see around" an object are aligned with the corners of its bounding box, where multiple faces intersect. The transformation from the initial to the final trajectory can be modeled as a process where trajectory segments proximal to salient objects differ via a 'perturbation field', causing them to deform towards the objects' periphery.

This trajectory provides a foundational scan of the room but lacks focus on specific foreground objects. To ensure the visual integrity of salient objects, the camera should deviate from the base trajectory to observe objects from informative viewpoints. For each salient object, represented by its bounding box, we aim to observe it from angles that reveal its side structures. Simply pulling the camera towards the object center often leads to flat views of object faces. Consequently, a static or purely distance-based trajectory often fails to capture the comprehensive geometry of occluded objects. We propose a bi-directional parallax exploration strategy. This approach modulates the camera path using an azimuth-dependent S-curve profile, effectively inducing lateral observation shifts while maintaining the global trajectory structure. For each anchor point $\mathbf{p}_l$ with azimuth angle $\phi_l$, we identify the nearest salient object $o^*$ and its centroid projected onto the horizontal plane, denoted as $\mathbf{c}_{o^*}$. Let $\phi_{o^*}$ be the azimuth of the object centroid. We define the relative azimuthal deviation $\Delta \phi_l$ as the normalized angular difference:
\begin{equation}
    \Delta \phi_l = \text{norm}(\phi_l - \phi_{o^*}) \in (-\pi, \pi],
\end{equation}
where $\operatorname{norm}(\cdot)$ wraps the angle to the principal interval. This relative angle serves as the control variable for our exploration logic: $\Delta \phi_l < 0$ indicates the camera is approaching the object, while $\Delta \phi_l > 0$ indicates departure. 

To achieve the desired ``glance-and-return" motion that inspects both lateral sides of the object, we construct a modulation function $W(\Delta \phi_l)$ based on the first derivative of the Gaussian function. Unlike a standard Gaussian which produces a unidirectional bulge, the first derivative provides a smooth and anti-symmetric S-shaped profile that passes through zero when the camera directly faces the object ($\Delta \phi_l \approx 0$):
\begin{equation}
    W(\Delta \phi_l) = \Delta \phi_l \cdot \exp\left(-\frac{\Delta \phi_l^2}{2\sigma_{\phi}(S_{o^*})^2}\right),
\end{equation}
where $S_{o^*}$ denotes the projected diagonal scale of the object. We dynamically condition the oscillation parameters on the object's 3D bounding box. We define the amplitude $\lambda(S_{o^*})$ and angular scope $\sigma_{\phi}(S_{o^*})$ as adaptive functions proportional to $S_{o^*}$, ensuring scale-consistent roaming that avoids collisions while maximizing parallax. The deformation is applied along the tangential vector $\mathbf{u}_l$, which is orthogonal to both the view vector $\mathbf{v}_l = \mathbf{c}_{o^*} - \mathbf{p}_l$ and the world up-axis $\mathbf{e}_y$:
\begin{equation}
    \mathbf{u}_l = \frac{\mathbf{v}_l \times \mathbf{e}_y}{||\mathbf{v}_l \times \mathbf{e}_y||_2}, \quad \tilde{\mathbf{p}}_l = \mathbf{p}_l - \lambda(S_{o^*}) \cdot W(\Delta \phi_l) \cdot \mathbf{u}_l.
\end{equation}

By applying this transformation, the adapted set $\{\tilde{\mathbf{p}}\}_{l=1}^L$ forms a wavy path that naturally swings outward to reveal occluded textures on the object's flanks and converges back to the base trajectory in other regions. Finally, we treat $\{\tilde{\mathbf{p}}\}_{l=1}^L$ as control knots and fit a closed periodic cubic spline. This yields a closed continuous trajectory $Tr$, which is generated adaptively according to the disentangled salient objects. Building upon this trajectory, we introduce perturbations to the cameras nearest to the salient objects, thereby generating additional views to further enhance the spatial coverage of the scene.

\subsection{Motion-injected RGBD Panorama Inpainting}
To achieve spatially consistent and visually plausible scene completion under dynamic viewpoints, we fine-tune the powerful image generation model \cite{rombach2022high} and introduce a Motion-injected RGBD Panorama Inpainting module. Unlike conventional inpainting models that treat missing regions as static image holes and train the models with random masks, our method explicitly incorporates camera motion cues to guide the inpainting process. This design enhances the inpainting model's ability to recover missing regions in novel views during camera movement, thereby ensuring better compatibility with our scene generation framework.

\subsubsection{Dataset Construction}
We leverage the Habitat \cite{habitat} simulator to curate a large-scale dataset from the Matterport3D environment containing pairs of panoramas $(\mathbf{I}_s, \mathbf{I}_t)$ captured under known camera motion $\Delta \mathbf{M}_{s\rightarrow t} = (\Delta \mathbf{R}, \Delta \mathbf{T})$. Specifically, we first acquire a comprehensive set of panoramic images along camera trajectories, which subsequently serves as the foundation for constructing the training data pairs.

Leveraging Habitat's autonomous navigation, we generate 30 distinct trajectories for each scene. To facilitate comprehensive data collection, a mobile agent is configured with a multi-sensor rig consisting of six RGBD cameras oriented in the cardinal directions (front, back, left, right, up, and down). For each sampled trajectory, we establish 50 recording points to capture camera poses and their corresponding six-view RGBD streams, which collectively form a cube map representation. These cube maps are subsequently projected into equirectangular panoramas. To ensure the quality of the dataset, a rigorous filtering process is applied: samples containing more than 2\% invalid (black) pixel areas were discarded. This refinement process yields a high-quality dataset comprising 49,987 annotated panoramic frames. Building upon the curated dataset, we synthesize training pairs by associating each anchor frame within a trajectory with its six nearest neighboring frames. For each pair, the source RGBD panorama and its corresponding camera parameters are utilized to reconstruct a 3D mesh proxy of the scene. This mesh is then projected onto the target pose to render a partially observed panorama $\mathbf{I}_r$, accompanied by a binary mask indicating disoccluded or missing regions. To ensure invariant scene scale, camera coordinates are normalized relative to the scene’s vertical extent (floor-to-ceiling distance). Finally, the constructed input data is $\mathbf{I}_r$, the associated mask, and the relative motion $\Delta \mathbf{M}_{s\rightarrow t}$, while the target panorama $\mathbf{I}_t$ serves as the ground truth.

\subsubsection{Model Architecture}
\begin{figure}[t]
\centering
\includegraphics[width=0.999\columnwidth]{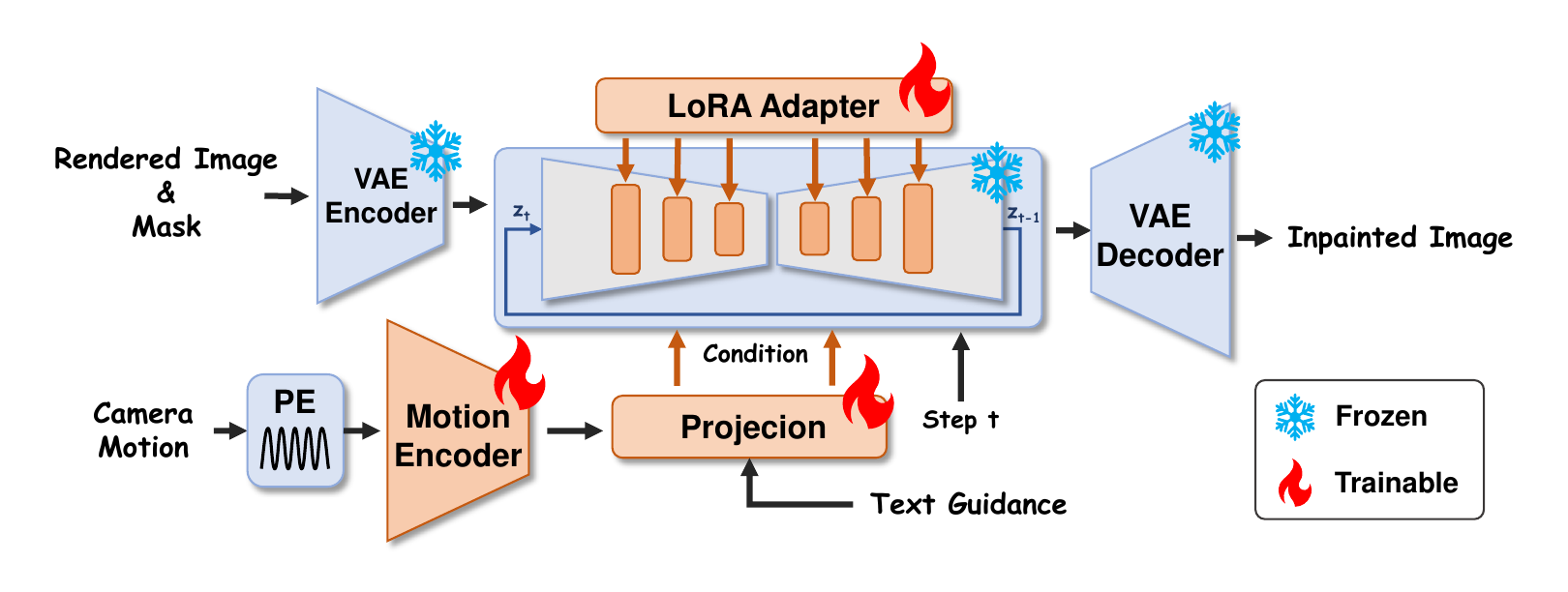} 
\caption{The architecture of the proposed motion-injected panorama inpainting model. We design a camera motion encoder to inject movement into the UNet. To train the interaction between static and dynamic features, we use LoRA adapters to fine-tune all the self-attention and cross-attention layers.}
\label{figure:inpainting}
\end{figure}
% The proposed Motion-Injected Inpainting model is trained in the latent space for both efficiency and high-frequency fidelity. 
As shown in Fig. \ref{figure:inpainting}, the core network is a UNet $\epsilon_\theta$. Concretely, we adopt the pretrained VAE encoder to map $\mathbf{I}_r$ and $\mathbf{I}_t$ into latent codes $\mathbf{z}_r$ and $\mathbf{z}_0$. At the denoising step $\tau$, the UNet takes the concatenation of the noised latent, the mask, and the latent of $\mathbf{I}_r$ as input, i.e., $\mathbf{x}_\tau = [\mathbf{z}_\tau, \mathbf{m}, \mathbf{z}_r]$ along the channel dimension. This design matches the pretrained inpainting interface and allows the model to simultaneously reason about the current noisy state, the spatial support of unknown regions, and the observed context. To explicitly inject motion cues, we condition the UNet on a compact representation of camera motion $\Delta \mathbf{M}_{s\rightarrow t}$. For each training pair, we compute the relative translation $\Delta \mathbf{T}$ between source and target camera centers and normalize it by an estimate of the scene vertical extent to remove scale ambiguity. To capture high-frequency movement details, we first apply sinusoidal positional encoding $\beta(\cdot)$ to the camera motion and then feed it into a Multi-Layer Perceptron (MLP):
\begin{equation}
\mathbf{z}_{cam} = \operatorname{MLP}(\beta(\Delta \mathbf{M}_{s\rightarrow t}))+\mathbf{o},
\end{equation}
where $\mathbf{z}_{cam}$ denotes the high-dimensional embedding of the camera motion, while $\mathbf{o}$ is a learnable pose token offset to absorb residual biases and improve conditioning capacity. We inject this embedding $\mathbf{z}_{cam}$ into the text-conditioning pathway via an additive ``expand'' strategy: for a text embedding sequence produced by the frozen CLIP text encoder, we broadcast the motion embedding $\mathbf{z}_{cam}$ across the sequence.

We fine-tune the diffusion model using Low-Rank Adaptation (LoRA) to retain the strong prior of the pretrained inpainting model. The original weights are frozen, and we attach LoRA adapters to the query, key, value, and output projections in the self-attention and cross-attention blocks. This confines optimization to a small set of low-rank matrices together with the motion encoder MLP (and also the learnable pose tokens), enabling stable training with limited compute. We further employ memory-efficient attention when available and mixed-precision training with upcasting of trainable parameters to full precision to maintain numerical robustness.

% Training follows the standard denoising diffusion objective. We sample a timestep $\tau$ uniformly, draw i.i.d.\ Gaussian noise $\boldsymbol{\epsilon}$, and obtain a noised latent $\mathbf{z}_\tau$ using the forward diffusion process implemented by the scheduler. The motion-injected UNet predicts the noise residual $\hat{\boldsymbol{\epsilon}}_\theta(\mathbf{x}_\tau, \tau,{\mathbf{z}_{cam}})$. In addition, we incorporate a teacher-student distillation term to regularize the fine-tuning: a frozen teacher UNet from the same pretrained checkpoint predicts $\hat{\boldsymbol{\epsilon}}$, and we penalize the discrepancy between teacher and student predictions. The overall objective is a weighted sum of the diffusion loss and the distillation loss, using a curriculum that emphasizes distillation in early iterations and transitions to stronger data-fitting later. 

The objective combines standard diffusion and teacher-student distillation. Given sampled timestep $\tau$, Gaussian noise $\boldsymbol{\epsilon}$, and noised latent $\mathbf{z}_\tau$, we minimize the reconstruction error of the motion-injected UNet prediction $\hat{\boldsymbol{\epsilon}}_\theta(\mathbf{x}_\tau, \tau,{\mathbf{z}_{cam}})$. Simultaneously, we regularize fine-tuning by penalizing the discrepancy between the student and a frozen teacher's prediction $\hat{\boldsymbol{\epsilon}}$. The objective is formalized as:
\begin{equation}
\begin{split}
\mathcal{L} = \quad & \lambda_1\mathbb{E}_{\mathbf{x}_\tau, \boldsymbol{\epsilon}, \tau, \mathbf{z}_{cam}} \left[ \|\boldsymbol{\epsilon} - \hat{\boldsymbol{\epsilon}}_\theta(\mathbf{x}_\tau, \tau, \mathbf{z}_{cam})\|_2^2 \right] \\
+ \, & \lambda_2\mathbb{E}_{\mathbf{x}_\tau, \hat{\boldsymbol{\epsilon}}, \tau, \mathbf{z}_{cam}} \left[ \|\hat{\boldsymbol{\epsilon}} - \hat{\boldsymbol{\epsilon}}_\theta(\mathbf{x}_\tau, \tau, \mathbf{z}_{cam})\|_2^2 \right].
\end{split}
\end{equation}

\subsubsection{Inference}
\begin{figure*}[t]
\centering
\includegraphics[width=0.999\linewidth]{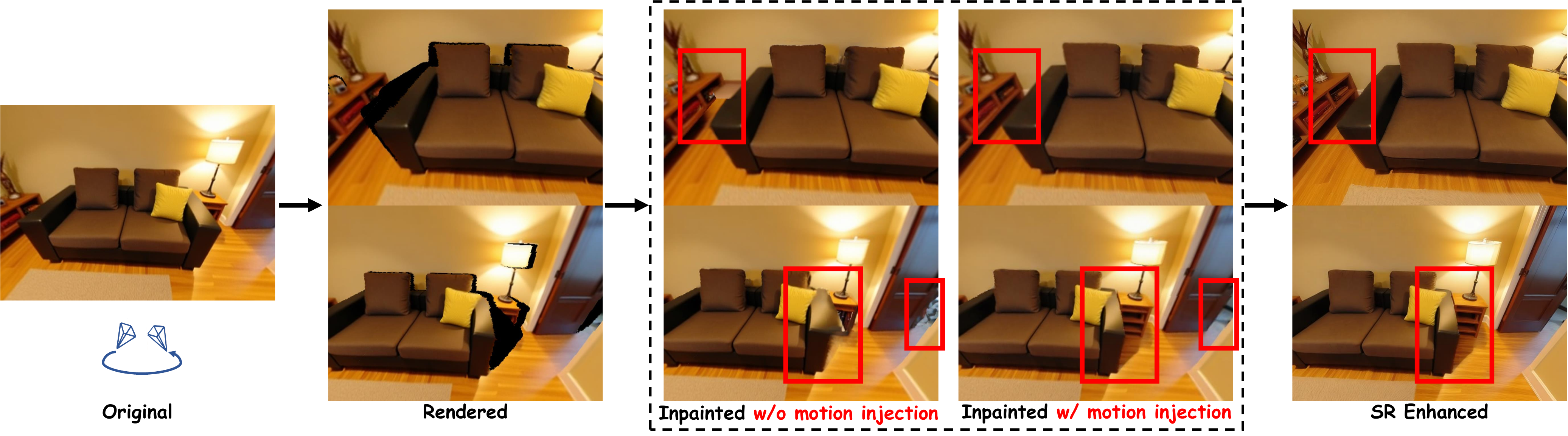} 
\caption{The visual examples of our inpainting and SR enhancement operation. The novel rendered views undergo inpainting and super-resolution to generate the final reference images. In the dashed box, we compare the inpainting results with and without motion injection.}
\label{figure:inpainting_example}
\end{figure*}
During the inference phase, the model takes the partially rendered panorama $I_{r}$ and the corresponding binary mask as input to synthesize the missing regions, explicitly conditioned on the encoded camera motion embedding $z_{cam}$ and textual guidance. To ensure high semantic fidelity relative to the local view, the text condition is dynamically derived from the Semantic Scene Graph constructed in the Scene Initialization stage. Specifically, we extract an object-centric subgraph $G_i$ focused on the salient object currently being explored and decode it into a brief text description. Upon generating the RGB content, we recover the geometry of the unknown regions utilizing the depth estimation and alignment methodology described in Sec. \ref{sec:initialization}. The visual examples are shown in Fig.\ref{figure:inpainting_example}. This visual evidence demonstrates that the model fine-tuned with camera motion yields superior performance. To guarantee geometric continuity, we introduce an additional constraint during this alignment process: the predicted depth scale and shift are optimized to ensure that the estimated depth within the known regions of $I_{r}$ tightly aligns with the existing depth values projected from the global mesh.

\subsection{Iterative Geometry Optimization and Refinement}
We adopt an iterative approach to progressively complete the 3D scene geometry. Starting with the initial panoramic view, our system follows the planned trajectory to sample novel viewpoints. At each step, we project the currently accumulated mesh into the new view frustum to identify unobserved regions. Our motion-injected inpainting model then synthesizes the missing RGB content, while depth and normal maps are predicted to recover the local geometry. To maintain global consistency, we unproject these new observations into 3D space and apply a geometric consistency check: points that conflict with the existing surface are discarded, while valid geometry is fused into the global mesh. This cycle of rendering, inpainting, and fusion allows us to incrementally extend the scene boundaries, transforming sparse initial observations into a dense and coherent surface representation.

While the consolidated mesh captures the coarse scene structure, it is often insufficient for photorealistic rendering due to discretization artifacts. To enhance visual fidelity, we employ 3D Gaussian Splatting (3DGS) for the final scene representation. We initialize the 3D Gaussians using the vertices and colors from our optimized mesh, which serves as a robust geometric prior to stabilize training. The optimization is then supervised by the set of synthesized panoramic views, which are decomposed into perspective cubemaps to provide multi-view constraints. By jointly optimizing the Gaussian parameters, we refine the scene's appearance to match the inpainted observations. This coarse-to-fine strategy effectively combines the structural reliability of explicit meshes with the rendering quality of volumetric representations, ensuring both geometric accuracy and photorealistic details.

\section{Experiments}
\begin{figure*}[t]
\centering
\includegraphics[width=0.99\textwidth]{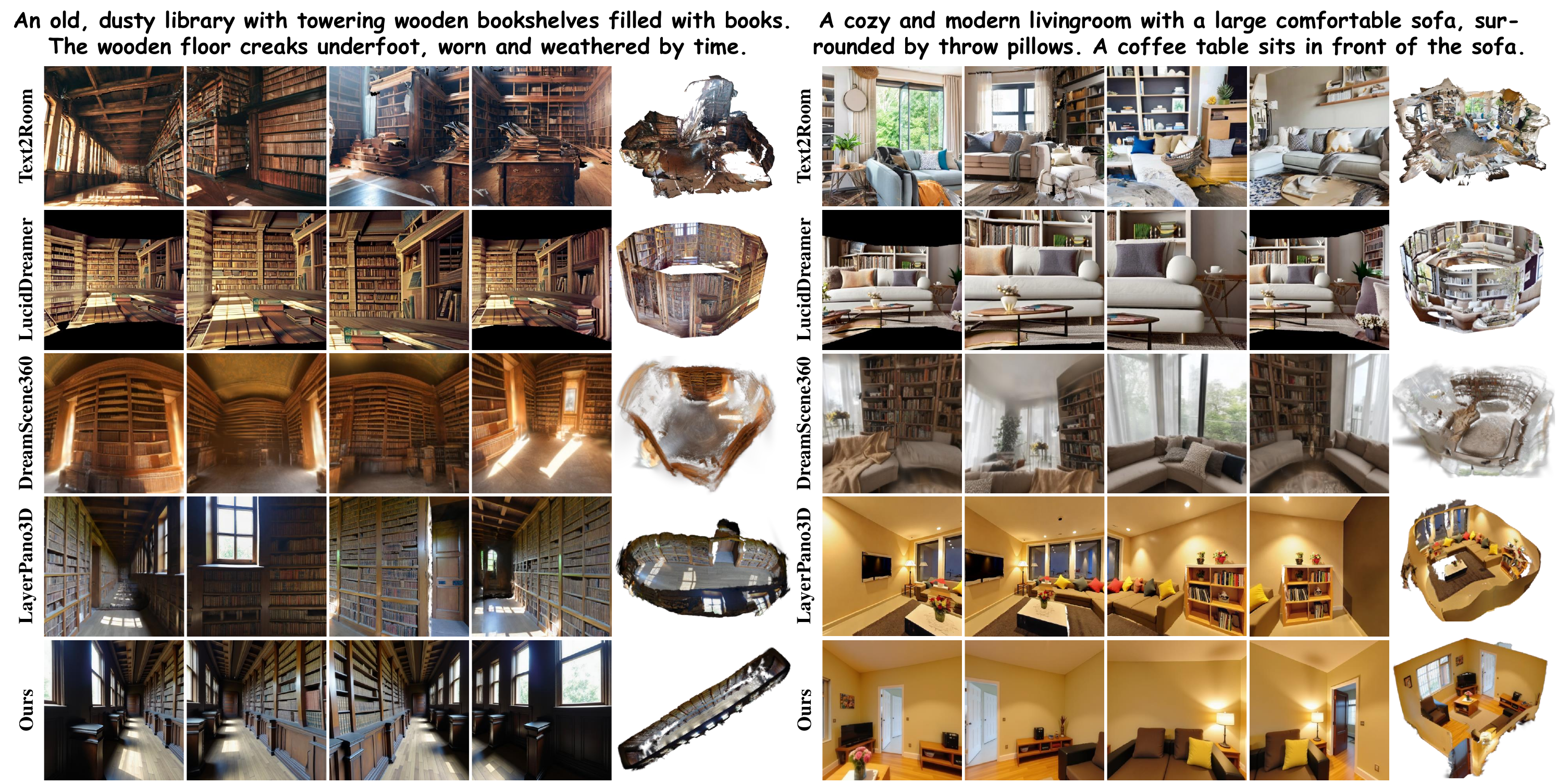}
\caption{Text-to-3D visual results of indoor scenes. Compared to other methods, ours generates indoor scenes of higher quality. Our rendered views exhibit greater content completeness, with object boundaries free from blurring or hole artifacts. Our 3D scenes exhibit more accurate geometry and crisp edges.}
\label{figure:t2s_in}
\end{figure*}
We conduct comprehensive evaluations, including quantitative and qualitative comparisons with state-of-the-art approaches, as well as ablation studies to investigate the effectiveness of individual components.

\subsection{Implementation Details}
We employ the same panorama generation model following LayerPano3D \cite{yang2025layerpano3d}, which is finetuned from the advanced FLUX model \cite{flux2024}, to generate the initial $1024 \times 2048$ panorama. In panoramic depth estimation and alignment, we optimize a global SDF (Sphere Distance Field) parameterized by a multi-resolution HashGrid ($L=16, N_{\min}=16, N_{\max}=2048, T=2^{19}, F=2$) cascaded with an MLP. Joint optimization with per-view affine parameters uses Adam for 2000 iterations. Learning rates are set to $10^{-1}$ (alignment) and $10^{-2}$ (SDF), with loss weights $\lambda_{reg}=0.1$ and $\lambda_{n}=\lambda_{tv}=0.01$. We reconstruct the mesh on a $1024 \times 2048$ grid with a $0.05$ edge threshold. To generate the semantic scene graph, we employ OpenAI GPT-5, identifying object categories and salient structures. For 2D semantic segmentation, we integrate SAM\cite{kirillov2023segment} with CLIP\cite{radford2021learning}. We use CLIP to compute similarity scores between the text prompts (derived from the scene graph) and image crops. We set $\gamma=0.4$ for the base camera trajectory and sample $24$ new views along the adaptive trajectory. During inpainting, we project the panorama into 6 cubemap faces, each with a resolution of $512 \times 512$ and a FoV of $90^{\circ}$. We apply morphological erosion and dilation using an $9 \times 9$ elliptical kernel to the masks to ensure seamless blending between the inpainted regions and the original context. The 6-DoF camera motion is embedded with sinusoidal positional encoding and then encoded through a 3-layer MLP. 3D Gaussians are optimized for 5,000 iterations on $512 \times 512$ perspective cubemap faces derived from inpainted panoramas. The position learning rate exponentially decays from $1.6 \times 10^{-4}$ to $1.6 \times 10^{-6}$, while other rates are fixed: feature $2.5 \times 10^{-3}$, opacity $0.05$, scaling $0.005$, and rotation $0.001$. Adaptive densification and pruning are performed every 100 iterations with a position gradient threshold of $0.0002$. The training and inference are conducted on GPUs with at least 48GB of VRAM.

\subsection{Comparison Baselines}
% Our method first generates a panorama from a textual prompt and integrates multi-modal priors and semantic information to generate 3D scenes. 
To comprehensively evaluate the performance of RoamScene3D, we compare our method against several state-of-the-art 3D generation approaches.
% These baselines represent different technical paradigms, including mesh-based optimization, Gaussian Splatting, and panoramic synthesis. 
\textbf{Text2Room} \cite{hollein2023text2room} is an iterative optimization-based method that constructs textured 3D meshes from text prompts. \textbf{LucidDreamer} \cite{lucid} is a domain-free generation framework leveraging 3D Gaussian Splatting. It initializes the scene by unprojecting generated multiview images onto a point cloud and optimizes the representation to ensure visual quality. \textbf{DreamScene360}~\cite{zhou2024dreamscene360} is a text-to-3D method specifically designed for panoramic scene generation. It generates an initial panorama to initialize 3D Gaussian Splatting and employs random camera perturbations near the origin to densify the scene representation. \textbf{LayerPano3D}~\cite{yang2025layerpano3d} is a framework for hyper-immersive scene generation that addresses occlusion by decomposing the scene into layers based on depth. We include them as baselines to evaluate the quality of text-to-3D generation. \textbf{PERF}~\cite{Wang_2024_perf} is a method that generates Panoramic Neural Radiance Fields from a single input panorama. \textbf{Pano2Room}~\cite{pu2024pano2room} is an approach that synthesizes novel views from a single indoor panorama using panoramic depth estimation and RGBD inpainting. We include them as baselines to evaluate the quality of image-to-3D generation. Among them, \textbf{LucidDreamer} supports both text-to-3D and image-to-3D generation.

\subsection{Qualitative Comparisons}

\begin{figure*}[t]
\centering
\includegraphics[width=0.99\textwidth]{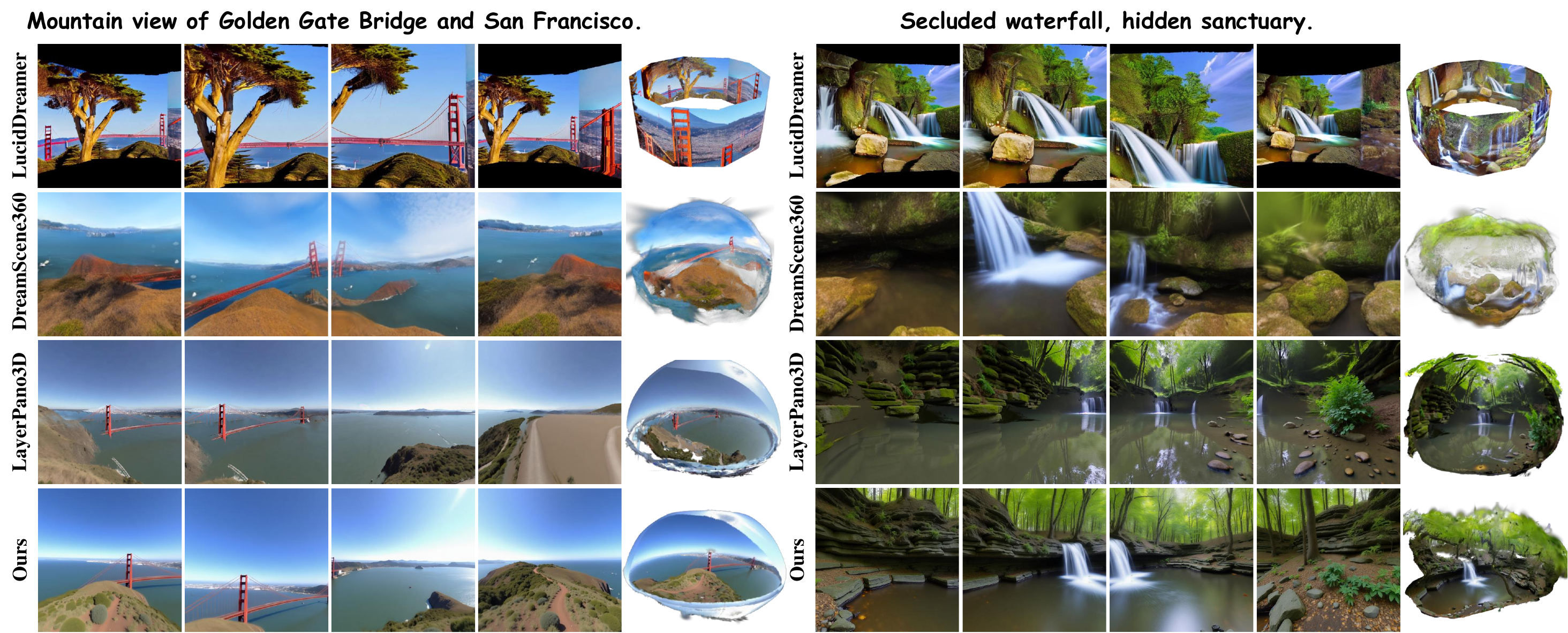}
\caption{Text-to-3D visual results of outdoor scenes. Our method generates more plausible content and achieves superior rendering quality. Furthermore, the estimated scene boundaries align more closely with real-world distributions, rather than modeling every scene as a sphere.}
\label{figure:t2s_out}
\end{figure*}

\begin{figure*}[t]
\centering
\includegraphics[width=0.99\textwidth]{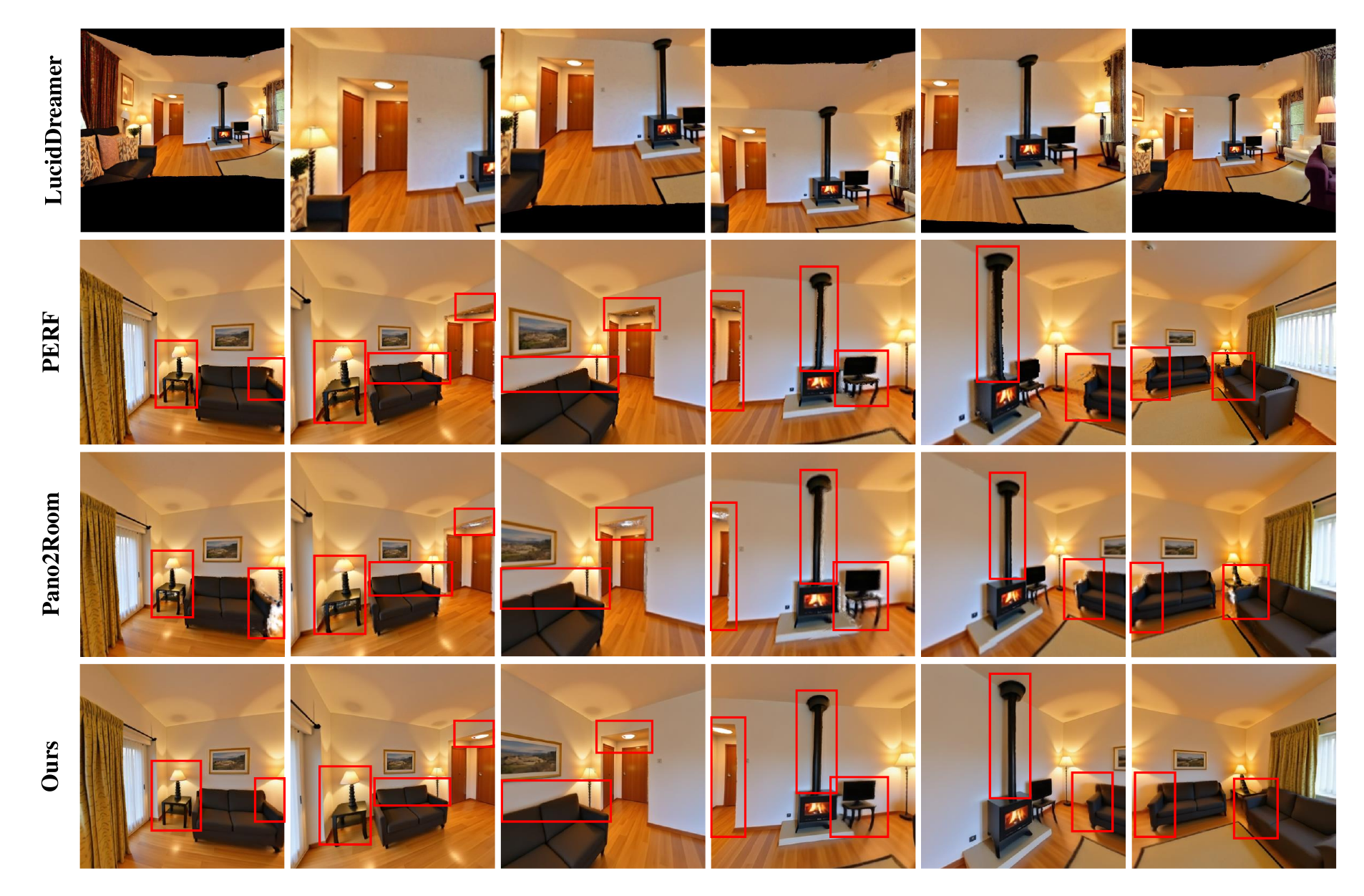}
\caption{Visual comparison with Image-to-3D methods. We utilize the panorama generated by our method during the initialization phase as input for competitive Image-to-3D approaches. Our generated scenes exhibit superior quality, characterized by greater object completeness and sharper edge definition.}
\label{figure:I2s}
\end{figure*}
We visualize the qualitative comparisons of both indoor scenes (in Fig.\ref{figure:t2s_in}) and outdoor scenes (in Fig.\ref{figure:t2s_out}) on competitive text-to-3D methods. We provide rendered views under diverse camera poses and 3D spatial views of the generated scenes. Leveraging perspective views, Text2Room and LucidDreamer generate a scene representation where valid content is populated exclusively within defined areas. DreamScene360, LayerPano3D, and our method RoamScene3D generate $360^\circ$ scenes, with outer boundaries partially removed to better visualize the internal structure. Our results show that RoamSceneD can generate immersive $360^\circ$ scenes for diverse indoor and outdoor settings conditioned on open-ended text prompts. Crucially, it outperforms existing methods, delivering superior fidelity in both 3D geometry and visual rendering. Besides, we feed the initial panorama generated by RoamScene3D into state-of-the-art image-to-3D generation methods, such as PERF and Pano2Room, and render images of the scenes from similar viewpoints. A comparison of the rendering results is presented in Fig.\ref{figure:I2s}. Our method exhibits sharper boundaries at object edges and more plausible spatial structures.

For the indoor scenes shown in Fig.\ref{figure:t2s_in}, the rendered images in each row are obtained by positioning the camera at different viewpoints and orientations. Text2Room produces a mesh representation characterized by severe geometric distortions, artifacts, and topological holes. The room edges are highly jagged, while the rendered views are compromised by significant blurring and fragmentation. Even with the 'rotate360' mode enabled for holistic scene coverage, LucidDreamer invariably yields scenes with an artifactual ring-shaped geometry, struggling to produce physically plausible spatial topology. DreamScene360 merely applies camera perturbations near the coordinate origin. Despite using panoramic images to optimize panoramic Gaussians, the generated scenes remain highly sparse, suffering from large holes and significant artifacts. The rendered images exhibit blurriness. LayerPano3D further decomposes the scene into layers, yielding more comprehensive scene coverage. However, due to the design of its layer-wise optimization algorithm, it tends to model the scene with a spherical geometry. For instance, the corridor in the left example is reconstructed as an ellipsoid, while the room walls on the right exhibit excessive curvature. Consequently, this spherical bias forces peripheral objects to be projected as flat surfaces onto the spherical shell, rather than retaining their 3D geometry. The rendered images also exhibit white holes. In contrast, our RoamScene3D demonstrates superior geometric generation capabilities in indoor environments. Specifically, it reconstructs the correct spatial distribution for the corridor on the left and maintains accurate wall boundaries for the room on the right. The generated scenes exhibit minimal voids, distortions, or artifacts, yielding high-fidelity rendered images that are perfectly consistent with the underlying 3D geometry.

For the outdoor scenes shown in Fig.\ref{figure:t2s_in}, the rendered images in each row are obtained by positioning the camera at different viewpoints and orientations. LucidDreamer continues to model the scene as a cylinder, resulting in a lack of content at the top and bottom. The scenes generated by DreamScene360 are sparse and contain numerous holes and artifacts. The scenes produced by them lack photorealism. In the left scene, the bridge generated by LayerPano3D is incomplete and exhibits blurring and distortions, while the generated 3D scene contains many holes. In the right scene, the waterfall lacks realism (see the 2nd and 3rd images from the right in the penultimate row), and rock structures are missing (see the 1st image from the right in the penultimate row). While modeling the scene on the left as spherical is justifiable within these two outdoor contexts, the scene on the right clearly deviates from a spherical geometry. Both DreamScene360 and LayerPano3D represent these two scenarios as spheres, which results in significant geometric distortions. This is particularly evident in the scene generated by LayerPano3D (right), where the trees and rocks along the pond are unnaturally projected onto a spherical surface, failing to maintain a realistic geometric structure. In contrast, our proposed RoamScene3D effectively addresses these aforementioned limitations. Scenes generated by our RoamScene3D are free from geometric incompleteness and distortions, aligning closely with real-world distributions, demonstrating superior performance over existing methods.

For the comparisons with image-to-3D methods shown in Fig.\ref{figure:I2s}, the rendered images in each row are obtained by positioning the camera at different viewpoints and orientations. LucidDreamer fails to generate full $360^{\circ} \times 180^{\circ}$ panoramic scenes, resulting in incomplete rendered images with black border artifacts. For PERF, Pano2Room, and our RoamScene3D, the generated scenes vary in scale and coordinate orientation, making it difficult to align them to a unified viewpoint. Therefore, we manually adjust the camera poses in each scene to be as similar as possible. The resulting renderings from these aligned perspectives are shown in each column. Due to the lack of adaptively generated object-aware camera trajectories, the images rendered by PERF and Pano2Room exhibit distortions or holes at the boundaries. Through visual comparison, our rendered images appear clearer, exhibiting sharper object boundaries and more coherent spatial structures. We highlight these key regions in red boxes.

\subsection{Quantitative Comparisons}
\begin{table*}[t]
\centering
\caption{Quantitative results on 60 different indoor and outdoor samples. The reported values are averaged over all test cases. Top three results: Red (bold), Orange (underlined), and Yellow.}
% \adjustbox{width=0.95\textwidth}{
\scalebox{1.00}{
\renewcommand{\arraystretch}{1.3}
\begin{tabular}{ccccccccc}
\toprule[1.2pt]
\multirow{2}{*}{\textbf{Methods}} & \multirow{2}{*}{\textbf{BRISQUE} {$\downarrow$}} & \multirow{2}{*}{\textbf{NIQE} {$\downarrow$}} & \multirow{2}{*}{\textbf{Inception-Score} {$\uparrow$}} & \multirow{2}{*}{\textbf{CLIP-Score} $\uparrow$} & \multicolumn{3}{c}{\textbf{CLIP-IQA} $\uparrow$} & \multirow{2}{*}{\textbf{Time (min)}} \\ \cmidrule{6-8}
 & & & & & \textbf{Quality} & \textbf{Colorful} & \textbf{Sharp} & \\ 
 \midrule
Text2Room \cite{hollein2023text2room} & 45.283 & 5.003 & 1.490 & 28.113 & 0.479 & 0.522 & 0.307 & 35.23 \\
LucidDreamer \cite{lucid} & 48.080 & 5.397 & 1.126 & 29.178 & 0.508 & 0.519 & 0.297 & 10.56 \\
PERF \cite{Wang_2024_perf} & 39.978 & 5.580 & \cellcolor{orange!30}\underline{1.642} & 30.050 & \cellcolor{yellow!50}{0.593} & 0.583 & \cellcolor{orange!30}\underline{0.367} & 56.09\\
Pano2Room \cite{pu2024pano2room} & \cellcolor{yellow!50}{38.667} & 5.445 & 1.484 & \cellcolor{orange!30}\underline{30.364} & 0.494 & \cellcolor{yellow!50}{0.592} & \cellcolor{yellow!50}{0.360} & 15.28 \\
DreamScene360 \cite{zhou2024dreamscene360} & 40.515 & \cellcolor{yellow!50}{5.412} & \cellcolor{yellow!50}{1.633} & 27.958 & 0.485 & 0.569 & 0.190 & 17.69 \\
LayerPano3D \cite{yang2025layerpano3d} & \cellcolor{orange!30}\underline{33.119} & \cellcolor{orange!30}\underline{4.182} & 1.530 & \cellcolor{yellow!50}{30.059} & \cellcolor{orange!30}\underline{0.607} & \cellcolor{orange!30}\underline{0.608} & 0.353 & 24.70\\
\textbf{RoamScene3D} \textbf{(Ours)} & \cellcolor{red!50}\textbf{29.563} & \cellcolor{red!50}\textbf{3.769} & \cellcolor{red!50}\textbf{1.645} & \cellcolor{red!50}\textbf{30.460} & \cellcolor{red!50}\textbf{0.665} & \cellcolor{red!50}\textbf{0.679} & \cellcolor{red!50}\textbf{0.453} & 23.08 \\
\bottomrule[1.2pt]
\end{tabular}
}
\label{table1}
\end{table*}
We conduct a comprehensive quantitative evaluation on 60 text prompts covering a diverse range of indoor and outdoor environments. Within the generated scenes, we generate 20 random rendering viewpoints distributed along a circular trajectory centered at the origin. The radius is defined as 0.3 times the nearest distance to the scene's horizontal boundary. Furthermore, we introduce minor perturbations along the vertical axis and randomize the viewing orientations for each camera. We compute these metrics for all rendered images across all scenes and report the averaged results in TABLE \ref{table1}.

To assess the generated 3D scenes from multiple dimensions, we employ seven standard metrics categorized into perceptual quality, semantic consistency, and aesthetic fidelity. Specifically, we utilize \textbf{BRISQUE} (Blind/Referenceless Image Spatial Quality Evaluator) and \textbf{NIQE} (Natural Image Quality Evaluator) as non-reference metrics to measure the naturalness and the presence of distortions in the rendered views, where lower scores indicate better visual quality. To evaluate the semantic alignment between the generated scenes and the input text, we report the \textbf{CLIP-Score}. Additionally, the \textbf{Inception-Score (IS)} is used to assess the diversity and meaningfulness of the generated content. Finally, to gauge the aesthetic appeal of the scenes, we employ the \textbf{CLIP-IQA} metric, specifically focusing on its \textbf{Quality}, \textbf{Colorful}, and \textbf{Sharp} sub-scores, where higher values correspond to superior aesthetic perception, color vibrancy, and image clarity, respectively.
\begin{figure}[]
\centering
\includegraphics[width=0.999\columnwidth]{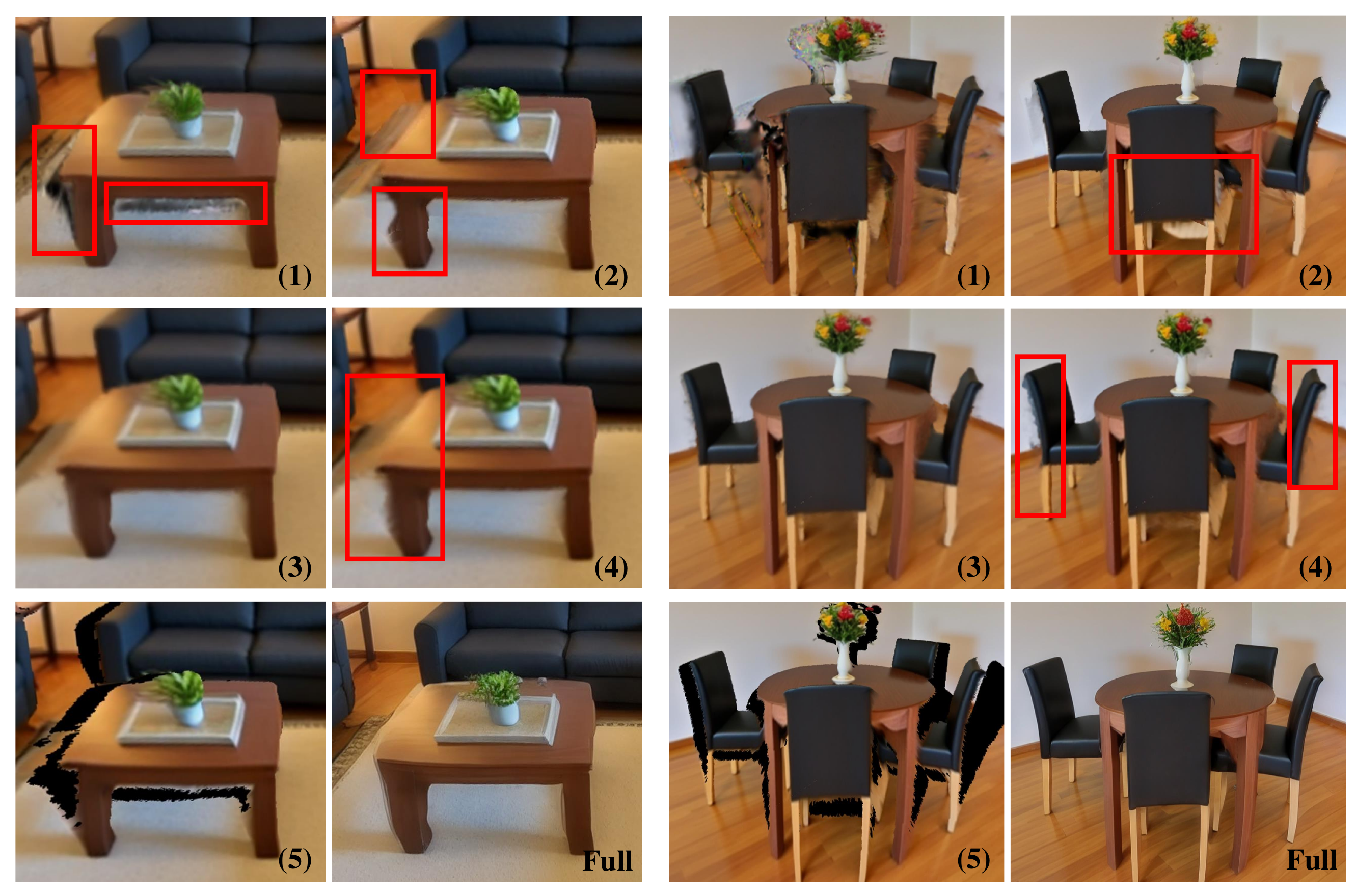} 
\caption{Qualitative visual results of the ablation study. Removing individual components leads to blurred object boundaries, geometric artifacts or holes, reduced texture sharpness, and semantic misalignment. The full RoamScene3D produces the most photorealistic, sharp, and spatially coherent scenes.}
\label{figure:ablation}
\end{figure}
As presented in TABLE \ref{table1}, our proposed RoamScene3D consistently outperforms state-of-the-art baselines across all evaluated metrics. In terms of perceptual quality, our method achieves the lowest scores, surpassing the second-best method by approximately 10.74\% in BRISQUE and 9.88\% in NIQE. This demonstrates a significant advantage over optimization-based methods like Text2Room and LucidDreamer, which often suffer from geometric artifacts and texture distortions. This superiority is largely attributed to our motion-injected inpainting module, which explicitly utilizes camera motion cues to maintain geometric consistency and eliminate visual artifacts during the generation of novel views. Furthermore, RoamScene3D demonstrates exceptional performance in semantic consistency and aesthetic quality. It outperforms the nearest competitors with improvements of 0.32\% in CLIP-Score and 0.18\% in Inception Score, validating the effectiveness of our semantic scene graph and adaptive roaming strategy in constructing scenes that are not only text-aligned but also structurally coherent with logical object placement. In the assessment of aesthetics, our method secures substantial gains over the competitors, achieving relative improvements of 9.56\%, 11.68\%, and 23.43\% in CLIP-IQA Quality, Colorfulness, and Sharpness, respectively. These results indicate that our approach, which integrates high-resolution panorama initialization with 3D Gaussian Splatting optimization, successfully preserves fine-grained details and produces vibrant, photorealistic environments that surpass the visual fidelity of existing panoramic generation methods like DreamScene360 and LayerPano3D. Regarding computational efficiency, our method requires approximately 23 minutes, achieving a favorable trade-off between high-fidelity rendering and inference speed compared to other optimization-based baselines like Text2Room, PERF, and LayerPano3D.

\begin{table}[t]
\centering
\caption{Quantitative results of ablation study. The metric names in the table are abbreviations.}
% \adjustbox{width=0.95\textwidth}{
\scalebox{1.00}{
\renewcommand{\arraystretch}{1.3}
\begin{tabular}{ccccccccc}
\toprule[1.2pt]
\textbf{Settings} & \textbf{BR.} $\downarrow$ & \textbf{NI.} $\downarrow$ & \textbf{IS} $\uparrow$ & \textbf{CS} $\uparrow$ & \textbf{Q.} $\uparrow$ & \textbf{C.} $\uparrow$ & \textbf{S.} $\uparrow$\\
 \midrule
\textbf{(1)} & 32.90 & 4.15 & 1.54 & 30.11 & 0.56 & 0.59 & 0.35 \\
\textbf{(2)} & 31.43 & 3.97 & 1.59 & 30.28 & 0.61 & 0.64 & 0.40 \\
\textbf{(3)} & 30.02 & 3.82 & 1.64 & 30.43 & 0.60 & 0.61 & 0.43 \\
\textbf{(4)} & 29.81 & 3.80 & 1.61 & 29.85 & 0.66 & 0.66 & 0.44 \\
\textbf{(5)} & 33.74 & 4.26 & 1.53 & 30.01 & 0.57 & 0.60 & 0.39 \\
\textbf{Full} & \textbf{29.56} & \textbf{3.77} & \textbf{1.65} & \textbf{30.46} & \textbf{0.67} & \textbf{0.68} & \textbf{0.45} \\
\bottomrule[1.2pt]
\end{tabular}
}
\label{table2}
\end{table}

\subsection{Ablation Study}

We conducted ablation studies to evaluate the key components of RoamScene3D. The quantitative results are reported in TABLE \ref{table2}, while the qualitative visual results are illustrated in Figure \ref{figure:ablation}. In TABLE \ref{table2}, the metric abbreviations BR., NI., IS, CS, Q., C., and S. stand for BRISQUE, NIQE, Inception-Score, CLIP-Score, Quality, Colorful, and Sharp, respectively. 
The settings for the ablation studies of (1), (2), (3), (4), and (5) are detailed respectively as follows:
\begin{itemize}
    \item \textbf{(1) w/o Adaptive Trajectory:} We replace the object-aware adaptive roaming trajectory with a fixed, collision-free base trajectory without perturbations.
    \item \textbf{(2) w/o Motion Injection:} The panorama inpainting model is replaced with a standard fine-tuned inpainting model that does not receive the camera motion vector $\Delta M$ as a condition.
    \item \textbf{(3) w/o PanoSR:} The super-resolution module is removed, directly using the output of the inpainting model for scene construction.
    \item \textbf{(4) w/o Object Disentanglement:} We remove the semantic scene graph and salient object disentanglement. The inpainting text guidance rely solely on global scene information rather than object-centric reasoning.
    \item \textbf{(5) w/o 3DGS Optimization:} The final 3D Gaussian Splatting optimization is omitted. The scene is represented and rendered using the optimized mesh obtained from the iterative fusion process.
\end{itemize}

First, the removal of the Adaptive Trajectory in \textbf{(1)} leads to a significant degradation in perceptual quality, with BRISQUE increasing from 29.56 to 32.90. As shown in Fig. 9, without the "glance-and-return" strategy, the camera fails to capture the lateral textures of salient objects, resulting in blurred artifacts when these objects are viewed from novel angles.
Second, the absence of Motion Injection in \textbf{(2)} causes a decline in geometric consistency. Without explicit motion cues, the inpainting model struggles to predict plausible disocclusions that align with large parallax changes, leading to a higher NIQE score.
Third, removing PanoSR in \textbf{(3)} primarily affects image sharpness and detail. While the structural metrics remain comparable, the Aesthetic Sharpness score drops from 0.45 to 0.43, resulting in scenes that lack high-frequency texture details.
Fourth, the impact of Object Disentanglement in \textbf{(4)} is most evident in semantic consistency. As indicated in Table II, this setting yields the lowest CLIP-Score (29.85). Without object-centric reasoning, the model lacks precise semantic guidance during inpainting, occasionally hallucinating incorrect textures on object surfaces.
Finally, relying solely on the Mesh representation in \textbf{(5)} results in the poorest visual quality across most metrics (e.g., worst BRISQUE). As visualized in Fig. 9, the mesh representation suffers from "flying pixel" artifacts and holes at depth discontinuities, whereas our full 3DGS model effectively fuses these imperfections to produce a continuous, photorealistic appearance. The full RoamScene3D framework achieves the best performance across all metrics, confirming that the synergy of these components is essential for high-fidelity 3D scene generation.

\section{Conclusions}
In this paper, we present RoamScene3D, a holistic framework for text-to-3D scene generation that effectively resolves the conflict between 2D generative priors and 3D spatial consistency. We identify that the primary bottleneck of existing methods lies in their lack of semantic spatial awareness, leading to geometric inconsistency and artifacts. To overcome this, We first combine the generation process with semantic reasoning. By leveraging a VLM-derived semantic scene graph, RoamScene3D understands the scene's layout and object relationships. This allows for Adaptive Trajectory Planning, which transforms scene exploration from a random process into an object-aware strategy to ensure comprehensive coverage of salient geometries. Second, we address the limitations of standard 2D inpainting in 3D contexts. Our Motion-injected Inpainting module models the correlation between camera movement and scene cclusions. This successfully optimizes 2D diffusion priors to handle complex occlusions. Quantitative and qualitative evaluations demonstrate that RoamScene3D produces scenes with superior photorealism and structural coherence. We believe this work offers a robust solution for bridging semantic guidance with high-fidelity 3D generation.

\bibliographystyle{IEEEtran}
\bibliography{IEEEabrv,references}

\begin{thebibliography}{10}
\providecommand{\url}[1]{#1}
\csname url@samestyle\endcsname
\providecommand{\newblock}{\relax}
\providecommand{\bibinfo}[2]{#2}
\providecommand{\BIBentrySTDinterwordspacing}{\spaceskip=0pt\relax}
\providecommand{\BIBentryALTinterwordstretchfactor}{4}
\providecommand{\BIBentryALTinterwordspacing}{\spaceskip=\fontdimen2\font plus
\BIBentryALTinterwordstretchfactor\fontdimen3\font minus \fontdimen4\font\relax}
\providecommand{\BIBforeignlanguage}[2]{{%
\expandafter\ifx\csname l@#1\endcsname\relax
\typeout{** WARNING: IEEEtran.bst: No hyphenation pattern has been}%
\typeout{** loaded for the language `#1'. Using the pattern for}%
\typeout{** the default language instead.}%
\else
\language=\csname l@#1\endcsname
\fi
#2}}
\providecommand{\BIBdecl}{\relax}
\BIBdecl

\bibitem{mildenhall2021nerf}
B.~Mildenhall, P.~P. Srinivasan, M.~Tancik, J.~T. Barron, R.~Ramamoorthi, and R.~Ng, ``Nerf: Representing scenes as neural radiance fields for view synthesis,'' \emph{Communications of the ACM}, vol.~65, no.~1, pp. 99--106, 2021.

\bibitem{3dgs}
B.~Kerbl, G.~Kopanas, T.~Leimk{\"u}hler, and G.~Drettakis, ``3d gaussian splatting for real-time radiance field rendering.'' \emph{ACM Trans. Graph.}, vol.~42, no.~4, pp. 139--1, 2023.

\bibitem{bautista2022gaudi}
M.~A. Bautista, P.~Guo, S.~Abnar, W.~Talbott, A.~Toshev, Z.~Chen, L.~Dinh, S.~Zhai, H.~Goh, D.~Ulbricht \emph{et~al.}, ``Gaudi: A neural architect for immersive 3d scene generation,'' \emph{Advances in Neural Information Processing Systems}, vol.~35, pp. 25\,102--25\,116, 2022.

\bibitem{chen2023scenedreamer}
Z.~Chen, G.~Wang, and Z.~Liu, ``Scenedreamer: Unbounded 3d scene generation from 2d image collections,'' \emph{IEEE transactions on pattern analysis and machine intelligence}, vol.~45, no.~12, pp. 15\,562--15\,576, 2023.

\bibitem{yu2025food3d}
D.~Yu, W.~Min, X.~Jin, Q.~Jiang, S.~Yao, and S.~Jiang, ``Food3d: Text-driven customizable 3d food generation with gaussian splatting,'' \emph{IEEE Transactions on Image Processing}, vol.~34, pp. 7290--7304, 2025.

\bibitem{zhu2025tg}
Y.~Zhu, Y.~Wang, and X.~Dong, ``Tg-tsgnet: A text-guided arbitrary-resolution terrain scene generation network,'' \emph{IEEE Transactions on Image Processing}, vol.~34, pp. 8614--8626, 2025.

\bibitem{li2026hyperbolic}
W.~Li, Y.~Lu, Y.~Chai, R.~Zhao, H.~Man, and X.~Fan, ``Hyperbolic hierarchical alignment reasoning network for text-3d retrieval,'' \emph{arXiv preprint arXiv:2511.11045}, 2025.

\bibitem{hollein2023text2room}
L.~H{\"o}llein, A.~Cao, A.~Owens, J.~Johnson, and M.~Nie{\ss}ner, ``Text2room: Extracting textured 3d meshes from 2d text-to-image models,'' in \emph{Proceedings of the IEEE/CVF International Conference on Computer Vision}, 2023, pp. 7909--7920.

\bibitem{fridman2023scenescape}
R.~Fridman, A.~Abecasis, Y.~Kasten, and T.~Dekel, ``Scenescape: Text-driven consistent scene generation,'' \emph{Advances in Neural Information Processing Systems}, vol.~36, pp. 39\,897--39\,914, 2023.

\bibitem{li2025riemann}
W.~Li, W.~Han, Y.~Chen, Y.~Chai, Y.~Lu, X.~Wang, and X.~Fan, ``Riemann-based multi-scale attention reasoning network for text-3d retrieval,'' in \emph{Proceedings of the AAAI Conference on Artificial Intelligence}, vol.~39, no.~17, 2025, pp. 18\,485--18\,493.

\bibitem{chu2025digging}
J.~Chu, W.~Li, X.~Wang, K.~Ning, Y.~Lu, and X.~Fan, ``Digging into intrinsic contextual information for high-fidelity 3d point cloud completion,'' in \emph{Proceedings of the AAAI Conference on Artificial Intelligence}, vol.~39, no.~3, 2025, pp. 2573--2581.

\bibitem{li2025hyperbolic}
W.~Li, Z.~Yang, W.~Han, H.~Man, X.~Wang, and X.~Fan, ``Hyperbolic-constraint point cloud reconstruction from single rgb-d images,'' in \emph{Proceedings of the AAAI Conference on Artificial Intelligence}, vol.~39, no.~5, 2025, pp. 4959--4967.

\bibitem{hu2024scenecraft}
Z.~Hu, A.~Iscen, A.~Jain, T.~Kipf, Y.~Yue, D.~A. Ross, C.~Schmid, and A.~Fathi, ``Scenecraft: An llm agent for synthesizing 3d scenes as blender code,'' in \emph{Forty-first International Conference on Machine Learning}, 2024.

\bibitem{yu2025wonderworld}
H.-X. Yu, H.~Duan, C.~Herrmann, W.~T. Freeman, and J.~Wu, ``Wonderworld: Interactive 3d scene generation from a single image,'' in \emph{Proceedings of the Computer Vision and Pattern Recognition Conference}, 2025, pp. 5916--5926.

\bibitem{nichol2021improved}
A.~Q. Nichol and P.~Dhariwal, ``Improved denoising diffusion probabilistic models,'' in \emph{International conference on machine learning}.\hskip 1em plus 0.5em minus 0.4em\relax PMLR, 2021, pp. 8162--8171.

\bibitem{rombach2022high}
R.~Rombach, A.~Blattmann, D.~Lorenz, P.~Esser, and B.~Ommer, ``High-resolution image synthesis with latent diffusion models,'' in \emph{Proceedings of the IEEE/CVF conference on computer vision and pattern recognition}, 2022, pp. 10\,684--10\,695.

\bibitem{lucid}
J.~Chung, S.~Lee, H.~Nam, J.~Lee, and K.~M. Lee, ``Luciddreamer: Domain-free generation of 3d gaussian splatting scenes,'' \emph{IEEE Transactions on Visualization and Computer Graphics}, vol.~31, no.~12, pp. 10\,640--10\,651, 2025.

\bibitem{li2024scenedreamer360}
W.~Li, F.~Cai, Y.~Mi, Z.~Yang, W.~Zuo, X.~Wang, and X.~Fan, ``Scenedreamer360: Text-driven 3d-consistent scene generation with panoramic gaussian splatting,'' \emph{arXiv preprint arXiv:2408.13711}, 2024.

\bibitem{zhou2024dreamscene360}
S.~Zhou, Z.~Fan, D.~Xu, H.~Chang, P.~Chari, T.~Bharadwaj, S.~You, Z.~Wang, and A.~Kadambi, ``Dreamscene360: Unconstrained text-to-3d scene generation with panoramic gaussian splatting,'' in \emph{European Conference on Computer Vision}.\hskip 1em plus 0.5em minus 0.4em\relax Springer, 2024, pp. 324--342.

\bibitem{pu2024pano2room}
G.~Pu, Y.~Zhao, and Z.~Lian, ``Pano2room: Novel view synthesis from a single indoor panorama,'' in \emph{SIGGRAPH Asia 2024 Conference Papers}, 2024, pp. 1--11.

\bibitem{xia2025scenepainter}
C.~Xia, S.~Zhang, F.~Liu, C.~Liu, K.~Hirunyaratsameewong, and Y.~Duan, ``Scenepainter: Semantically consistent perpetual 3d scene generation with concept relation alignment,'' in \emph{Proceedings of the IEEE/CVF International Conference on Computer Vision}, 2025, pp. 28\,808--28\,817.

\bibitem{yang2025layerpano3d}
S.~Yang, J.~Tan, M.~Zhang, T.~Wu, G.~Wetzstein, Z.~Liu, and D.~Lin, ``Layerpano3d: Layered 3d panorama for hyper-immersive scene generation,'' in \emph{Proceedings of the special interest group on computer graphics and interactive techniques conference conference papers}, 2025, pp. 1--10.

\bibitem{Kim_2023_CVPR}
S.~W. Kim, B.~Brown, K.~Yin, K.~Kreis, K.~Schwarz, D.~Li, R.~Rombach, A.~Torralba, and S.~Fidler, ``Neuralfield-ldm: Scene generation with hierarchical latent diffusion models,'' in \emph{Proceedings of the IEEE/CVF conference on computer vision and pattern recognition}, 2023, pp. 8496--8506.

\bibitem{liu2023pyramid}
Y.~Liu, X.~Li, X.~Li, L.~Qi, C.~Li, and M.-H. Yang, ``Pyramid diffusion for fine 3d large scene generation,'' \emph{arXiv preprint arXiv:2311.12085}, 2023.

\bibitem{wu2024blockfusion}
Z.~Wu, Y.~Li, H.~Yan, T.~Shang, W.~Sun, S.~Wang, R.~Cui, W.~Liu, H.~Sato, H.~Li \emph{et~al.}, ``Blockfusion: Expandable 3d scene generation using latent tri-plane extrapolation,'' \emph{ACM Transactions on Graphics (ToG)}, vol.~43, no.~4, pp. 1--17, 2024.

\bibitem{meng2025lt3sd}
Q.~Meng, L.~Li, M.~Nie{\ss}ner, and A.~Dai, ``Lt3sd: Latent trees for 3d scene diffusion,'' in \emph{Proceedings of the Computer Vision and Pattern Recognition Conference}, 2025, pp. 650--660.

\bibitem{Ren_2022_CVPR}
X.~Ren and X.~Wang, ``Look outside the room: Synthesizing a consistent long-term 3d scene video from a single image,'' in \emph{Proceedings of the IEEE/CVF Conference on Computer Vision and Pattern Recognition}, 2022, pp. 3563--3573.

\bibitem{Li_2023_CVPR}
W.~Li, X.~Chen, J.~Wang, and B.~Chen, ``Patch-based 3d natural scene generation from a single example,'' in \emph{Proceedings of the IEEE/CVF Conference on Computer Vision and Pattern Recognition}, 2023, pp. 16\,762--16\,772.

\bibitem{zhang2024text2nerf}
J.~Zhang, X.~Li, Z.~Wan, C.~Wang, and J.~Liao, ``Text2nerf: Text-driven 3d scene generation with neural radiance fields,'' \emph{IEEE Transactions on Visualization and Computer Graphics}, vol.~30, no.~12, pp. 7749--7762, 2024.

\bibitem{Yu_2024_CVPR}
H.-X. Yu, H.~Duan, J.~Hur, K.~Sargent, M.~Rubinstein, W.~T. Freeman, F.~Cole, D.~Sun, N.~Snavely, J.~Wu \emph{et~al.}, ``Wonderjourney: Going from anywhere to everywhere,'' in \emph{Proceedings of the IEEE/CVF Conference on Computer Vision and Pattern Recognition}, 2024, pp. 6658--6667.

\bibitem{Wang_2024_perf}
G.~Wang, P.~Wang, Z.~Chen, W.~Wang, C.~C. Loy, and Z.~Liu, ``Perf: Panoramic neural radiance field from a single panorama,'' \emph{IEEE Transactions on Pattern Analysis and Machine Intelligence}, vol.~46, no.~10, pp. 6905--6918, 2024.

\bibitem{RGBD2}
J.~Lei, J.~Tang, and K.~Jia, ``Rgbd2: Generative scene synthesis via incremental view inpainting using rgbd diffusion models,'' in \emph{Proceedings of the IEEE/CVF conference on computer vision and pattern recognition}, 2023, pp. 8422--8434.

\bibitem{zhou2025recurrent}
Y.~Zhou, D.~Ye, H.~Zhang, X.~Xu, H.~Sun, Y.~Xu, X.~Liu, and Y.~Zhou, ``Recurrent diffusion for 3d point cloud generation from a single image,'' \emph{IEEE Transactions on Image Processing}, 2025.

\bibitem{xie2025tri}
Y.~Xie, H.~Xiao, and W.~Kang, ``Tri 2 plane: Advancing neural implicit surface reconstruction for indoor scenes,'' \emph{IEEE Transactions on Multimedia}, 2025.

\bibitem{zuo2025learning}
Y.~Zuo, Y.~Hu, Y.~Xu, Z.~Wang, Y.~Fang, J.~Yan, W.~Jiang, Y.~Peng, and Y.~Huang, ``Learning guided implicit depth function with scale-aware feature fusion,'' \emph{IEEE Transactions on Image Processing}, 2025.

\bibitem{li2025ustc}
Z.~Li, J.~Liao, C.~Tang, H.~Zhang, Y.~Li, Y.~Bian, X.~Sheng, X.~Feng, Y.~Li, C.~Gao \emph{et~al.}, ``Ustc-td: A test dataset and benchmark for image and video coding in 2020s,'' \emph{IEEE Transactions on Multimedia}, 2025.

\bibitem{chen2025MuTri}
Z.~Chen, H.~Wang, C.~Ou, and X.~Li, ``Mutri: Multi-view tri-alignment for oct to octa 3d image translation,'' in \emph{Proceedings of the IEEE/CVF Conference on Computer Vision and Pattern Recognition}, 2025.

\bibitem{Cai_2023_ICCV}
S.~Cai, E.~R. Chan, S.~Peng, M.~Shahbazi, A.~Obukhov, L.~Van~Gool, and G.~Wetzstein, ``Diffdreamer: Towards consistent unsupervised single-view scene extrapolation with conditional diffusion models,'' in \emph{Proceedings of the IEEE/CVF International Conference on Computer Vision}, 2023, pp. 2139--2150.

\bibitem{RoomDreamer}
L.~Song, L.~Cao, H.~Xu, K.~Kang, F.~Tang, J.~Yuan, and Y.~Zhao, ``Roomdreamer: Text-driven 3d indoor scene synthesis with coherent geometry and texture,'' \emph{arXiv preprint arXiv:2305.11337}, 2023.

\bibitem{zhangmonst3r}
J.~Zhang, C.~Herrmann, J.~Hur, V.~Jampani, T.~Darrell, F.~Cole, D.~Sun, and M.-H. Yang, ``Monst3r: A simple approach for estimating geometry in the presence of motion,'' in \emph{The Thirteenth International Conference on Learning Representations}, 2024.

\bibitem{zhang20243d}
S.~Zhang, Y.~Zhang, Q.~Zheng, R.~Ma, W.~Hua, H.~Bao, W.~Xu, and C.~Zou, ``3d-scenedreamer: Text-driven 3d-consistent scene generation,'' in \emph{Proceedings of the IEEE/CVF Conference on Computer Vision and Pattern Recognition}, 2024, pp. 10\,170--10\,180.

\bibitem{chen2025learning}
K.~Chen, Z.~Yuan, H.~Xiao, T.~Mao, and Z.~Wang, ``Learning multi-view stereo with geometry-aware prior,'' \emph{IEEE Transactions on Circuits and Systems for Video Technology}, 2025.

\bibitem{ju2025revisiting}
Y.~Ju, B.~Shi, B.~Wen, K.-M. Lam, X.~Jiang, and A.~C. Kot, ``Revisiting one-stage deep uncalibrated photometric stereo via fourier embedding,'' \emph{IEEE Transactions on Pattern Analysis and Machine Intelligence}, 2025.

\bibitem{xiao2025enhanced}
H.~Xiao, W.~Kang, Y.~Guo, H.~Liu, and Y.~He, ``Enhanced geometry and semantics for camera-based 3d semantic scene completion,'' \emph{IEEE Transactions on Image Processing}, vol.~35, pp. 1--13, 2025.

\bibitem{li2024art3d}
P.~Li, C.~Tang, Q.~Huang, and Z.~Li, ``Art3d: 3d gaussian splatting for text-guided artistic scenes generation,'' \emph{arXiv preprint arXiv:2405.10508}, 2024.

\bibitem{mei2024camera}
J.~Mei, Y.~Yang, M.~Wang, J.~Zhu, J.~Ra, Y.~Ma, L.~Li, and Y.~Liu, ``Camera-based 3d semantic scene completion with sparse guidance network,'' \emph{IEEE Transactions on Image Processing}, 2024.

\bibitem{wang2019planit}
K.~Wang, Y.-A. Lin, B.~Weissmann, M.~Savva, A.~X. Chang, and D.~Ritchie, ``Planit: Planning and instantiating indoor scenes with relation graph and spatial prior networks,'' \emph{ACM Transactions on Graphics (TOG)}, vol.~38, no.~4, pp. 1--15, 2019.

\bibitem{paschalidou2021atiss}
D.~Paschalidou, A.~Kar, M.~Shugrina, K.~Kreis, A.~Geiger, and S.~Fidler, ``Atiss: Autoregressive transformers for indoor scene synthesis,'' \emph{Advances in Neural Information Processing Systems}, vol.~34, pp. 12\,013--12\,026, 2021.

\bibitem{gao2024graphdreamer}
G.~Gao, W.~Liu, A.~Chen, A.~Geiger, and B.~Sch{\"o}lkopf, ``Graphdreamer: Compositional 3d scene synthesis from scene graphs,'' in \emph{Proceedings of the IEEE/CVF Conference on Computer Vision and Pattern Recognition}, 2024, pp. 21\,295--21\,304.

\bibitem{zhai2023commonscenes}
G.~Zhai, E.~P. {\"O}rnek, S.-C. Wu, Y.~Di, F.~Tombari, N.~Navab, and B.~Busam, ``Commonscenes: Generating commonsense 3d indoor scenes with scene graph diffusion,'' \emph{Advances in Neural Information Processing Systems}, vol.~36, pp. 30\,026--30\,038, 2023.

\bibitem{tang2024diffuscene}
J.~Tang, Y.~Nie, L.~Markhasin, A.~Dai, J.~Thies, and M.~Nie{\ss}ner, ``Diffuscene: Denoising diffusion models for generative indoor scene synthesis,'' in \emph{Proceedings of the IEEE/CVF conference on computer vision and pattern recognition}, 2024, pp. 20\,507--20\,518.

\bibitem{yang2024physcene}
Y.~Yang, B.~Jia, P.~Zhi, and S.~Huang, ``Physcene: Physically interactable 3d scene synthesis for embodied ai,'' in \emph{Proceedings of the IEEE/CVF Conference on Computer Vision and Pattern Recognition}, 2024, pp. 16\,262--16\,272.

\bibitem{fang2023ctrl}
C.~Fang, Y.~Dong, K.~Luo, X.~Hu, R.~Shrestha, and P.~Tan, ``Ctrl-room: Controllable text-to-3d room meshes generation with layout constraints,'' \emph{arXiv preprint arXiv:2310.03602}, 2023.

\bibitem{xu2023discoscene}
Y.~Xu, M.~Chai, Z.~Shi, S.~Peng, I.~Skorokhodov, A.~Siarohin, C.~Yang, Y.~Shen, H.-Y. Lee, B.~Zhou \emph{et~al.}, ``Discoscene: Spatially disentangled generative radiance fields for controllable 3d-aware scene synthesis,'' in \emph{Proceedings of the IEEE/CVF conference on computer vision and pattern recognition}, 2023, pp. 4402--4412.

\bibitem{zhou2024gala3d}
X.~Zhou, X.~Ran, Y.~Xiong, J.~He, Z.~Lin, Y.~Wang, D.~Sun, and M.-H. Yang, ``Gala3d: Towards text-to-3d complex scene generation via layout-guided generative gaussian splatting,'' \emph{arXiv preprint arXiv:2402.07207}, 2024.

\bibitem{merrell2011interactive}
P.~Merrell, E.~Schkufza, Z.~Li, M.~Agrawala, and V.~Koltun, ``Interactive furniture layout using interior design guidelines,'' \emph{ACM transactions on graphics (TOG)}, vol.~30, no.~4, pp. 1--10, 2011.

\bibitem{fisher2012example}
M.~Fisher, D.~Ritchie, M.~Savva, T.~Funkhouser, and P.~Hanrahan, ``Example-based synthesis of 3d object arrangements,'' \emph{ACM Transactions on Graphics (TOG)}, vol.~31, no.~6, pp. 1--11, 2012.

\bibitem{qi2018human}
S.~Qi, Y.~Zhu, S.~Huang, C.~Jiang, and S.-C. Zhu, ``Human-centric indoor scene synthesis using stochastic grammar,'' in \emph{Proceedings of the IEEE Conference on Computer Vision and Pattern Recognition}, 2018, pp. 5899--5908.

\bibitem{zhao2024roomdesigner}
Y.~Zhao, Z.~Zhao, J.~Li, S.~Dong, and S.~Gao, ``Roomdesigner: Encoding anchor-latents for style-consistent and shape-compatible indoor scene generation,'' in \emph{2024 International Conference on 3D Vision (3DV)}.\hskip 1em plus 0.5em minus 0.4em\relax IEEE, 2024, pp. 1413--1423.

\bibitem{Xie_2024_CVPR}
H.~Xie, Z.~Chen, F.~Hong, and Z.~Liu, ``Citydreamer: Compositional generative model of unbounded 3d cities,'' in \emph{Proceedings of the IEEE/CVF conference on computer vision and pattern recognition}, 2024, pp. 9666--9675.

\bibitem{li2025language}
W.~Li, W.~Han, H.~Man, W.~Zuo, X.~Fan, and Y.~Tian, ``Language-guided graph representation learning for video summarization,'' \emph{IEEE Transactions on Pattern Analysis and Machine Intelligence}, 2025.

\bibitem{InstructScene}
C.~Lin and Y.~Mu, ``Instructscene: Instruction-driven 3d indoor scene synthesis with semantic graph prior,'' in \emph{International Conference on Learning Representations (ICLR)}, 2024.

\bibitem{li2025spiking}
W.~Li, W.~Han, L.-J. Deng, R.~Xiong, and X.~Fan, ``Spiking variational graph representation inference for video summarization,'' \emph{IEEE Transactions on Image Processing}, 2025.

\bibitem{wei2023lego}
Q.~A. Wei, S.~Ding, J.~J. Park, R.~Sajnani, A.~Poulenard, S.~Sridhar, and L.~Guibas, ``Lego-net: Learning regular rearrangements of objects in rooms,'' in \emph{Proceedings of the IEEE/CVF Conference on Computer Vision and Pattern Recognition}, 2023, pp. 19\,037--19\,047.

\bibitem{zhang2024towards}
Q.~Zhang, C.~Wang, A.~Siarohin, P.~Zhuang, Y.~Xu, C.~Yang, D.~Lin, B.~Zhou, S.~Tulyakov, and H.-Y. Lee, ``Towards text-guided 3d scene composition,'' in \emph{Proceedings of the IEEE/CVF Conference on Computer Vision and Pattern Recognition}, 2024, pp. 6829--6838.

\bibitem{epstein2024disentangled}
D.~Epstein, B.~Poole, B.~Mildenhall, A.~A. Efros, and A.~Holynski, ``Disentangled 3d scene generation with layout learning,'' \emph{arXiv preprint arXiv:2402.16936}, 2024.

\bibitem{Wu_2023_ICCV}
Q.~Wu, K.~Wang, K.~Li, J.~Zheng, and J.~Cai, ``Objectsdf++: Improved object-compositional neural implicit surfaces,'' in \emph{Proceedings of the IEEE/CVF International Conference on Computer Vision}, 2023, pp. 21\,764--21\,774.

\bibitem{Bahmani_2023_ICCV}
S.~Bahmani, J.~J. Park, D.~Paschalidou, X.~Yan, G.~Wetzstein, L.~Guibas, and A.~Tagliasacchi, ``Cc3d: Layout-conditioned generation of compositional 3d scenes,'' in \emph{Proceedings of the IEEE/CVF International Conference on Computer Vision}, 2023, pp. 7171--7181.

\bibitem{bai2024360}
J.~Bai, L.~Huang, J.~Guo, W.~Gong, Y.~Li, and Y.~Guo, ``360-gs: Layout-guided panoramic gaussian splatting for indoor roaming,'' \emph{arXiv preprint arXiv:2402.00763}, 2024.

\bibitem{Schult_2024_CVPR}
J.~Schult, S.~Tsai, L.~H{\"o}llein, B.~Wu, J.~Wang, C.-Y. Ma, K.~Li, X.~Wang, F.~Wimbauer, Z.~He \emph{et~al.}, ``Controlroom3d: Room generation using semantic proxy rooms,'' in \emph{Proceedings of the IEEE/CVF Conference on Computer Vision and Pattern Recognition}, 2024, pp. 6201--6210.

\bibitem{jiao2025clip}
S.~Jiao, H.~Dong, Y.~Yin, Z.~Jie, Y.~Qian, Y.~Zhao, H.~Shi, and Y.~Wei, ``Clip-gs: Unifying vision-language representation with 3d gaussian splatting,'' in \emph{Proceedings of the IEEE/CVF International Conference on Computer Vision}, 2025, pp. 4670--4680.

\bibitem{mu2024learning}
T.-J. Mu, M.-Y. Shen, Y.-K. Lai, and S.-M. Hu, ``Learning virtual view selection for 3d scene semantic segmentation,'' \emph{IEEE Transactions on Image Processing}, 2024.

\bibitem{habitat}
X.~Puig, E.~Undersander, A.~Szot, M.~D. Cote, T.-Y. Yang, R.~Partsey, R.~Desai, A.~W. Clegg, M.~Hlavac, S.~Y. Min \emph{et~al.}, ``Habitat 3.0: A co-habitat for humans, avatars and robots,'' \emph{arXiv preprint arXiv:2310.13724}, 2023.

\bibitem{flux2024}
B.~F. Labs, ``Flux,'' \url{https://github.com/black-forest-labs/flux}, 2024.

\bibitem{kirillov2023segment}
A.~Kirillov, E.~Mintun, N.~Ravi, H.~Mao, C.~Rolland, L.~Gustafson, T.~Xiao, S.~Whitehead, A.~C. Berg, W.-Y. Lo \emph{et~al.}, ``Segment anything,'' in \emph{Proceedings of the IEEE/CVF international conference on computer vision}, 2023, pp. 4015--4026.

\bibitem{radford2021learning}
A.~Radford, J.~W. Kim, C.~Hallacy, A.~Ramesh, G.~Goh, S.~Agarwal, G.~Sastry, A.~Askell, P.~Mishkin, J.~Clark \emph{et~al.}, ``Learning transferable visual models from natural language supervision,'' in \emph{International conference on machine learning}.\hskip 1em plus 0.5em minus 0.4em\relax PmLR, 2021, pp. 8748--8763.

\end{thebibliography}
\clearpage

\end{document}